%% file: main.tex
\documentclass[10pt,twocolumn,letterpaper]{article}
\usepackage[accsupp]{axessibility}
\usepackage[pagenumbers]{cvpr} 

\input{preamble}

\makeatletter
\renewcommand{\maketag@@@}[1]{\hbox{\m@th\normalsize\normalfont#1}}%
\makeatother

\definecolor{cvprblue}{rgb}{0.21,0.49,0.74}

\usepackage{makecell}
\usepackage{multirow}
\usepackage{algorithm}
\usepackage{algpseudocode}

\def\paperID{8855} 
\def\confName{CVPR}
\def\confYear{2024}

\usepackage{textcomp}
\usepackage{xcolor}
\newcommand\Add[1]{#1}
\newcommand\Change[1]{#1}

\title{DiffCast: A Unified Framework via Residual Diffusion for Precipitation Nowcasting}

\author{Demin Yu\textsuperscript{1}, 
Xutao Li\thanks{Corresponding author} \textsuperscript{1}, 
Yunming Ye\textsuperscript{1}, Baoquan Zhang\textsuperscript{1}, Chuyao Luo\textsuperscript{1}, Kuai Dai\textsuperscript{1}, Rui Wang\textsuperscript{2}, Xunlai Chen\textsuperscript{2}\\
\textsuperscript{1}Harbin Institute of Technology, Shenzhen; \textsuperscript{2}Shenzhen Meteorological Bureau\\
{\tt\small deminy@stu.hit.edu.cn, \{lixutao, yeyunming, baoquanzhang\}@hit.edu.cn,}\\
{\tt\small luochuyao.dalian@gmail.com,
\{daikuai\_hit, wangr09, cxlxun\}@163.com}
}

\begin{document}
\maketitle
\input{sec/0_abstract}    
\input{sec/1_intro}

\input{sec/2_formatting}
{
    \small
    \bibliographystyle{ieeenat_fullname}
    \bibliography{main}
}

\input{sec/X_suppl}

\end{document}

%% file: preamble.tex
\usepackage[dvipsnames]{xcolor}


%% file: sec/0_abstract.tex
\begin{abstract}
Precipitation nowcasting is an important spatio-temporal prediction task to predict the radar echoes sequences based on current observations, which can serve both meteorological science and smart city applications. Due to the chaotic evolution nature of the precipitation systems, it is  a very challenging problem. Previous studies address the problem either from the perspectives of deterministic modeling or probabilistic modeling. However, their predictions suffer from the blurry, high-value echoes fading away and position inaccurate issues. The root reason of these issues is that the chaotic evolutionary precipitation systems are not appropriately modeled. Inspired by the nature of the systems, we propose to decompose and model them from the perspective of global deterministic motion and local stochastic variations with residual mechanism. A unified and flexible framework that can equip any type of spatio-temporal models is proposed based on residual diffusion, which effectively tackles the shortcomings of previous methods. Extensive experimental results on four publicly available radar datasets demonstrate the effectiveness and superiority of the proposed framework, compared to state-of-the-art techniques. Our code is publicly available at https://github.com/DeminYu98/DiffCast. 

\end{abstract}


%% file: sec/1_intro.tex
\section{Introduction}
\label{sec:intro}

\begin{figure}
    \centering
    \includegraphics[width = 1.0\columnwidth, ]{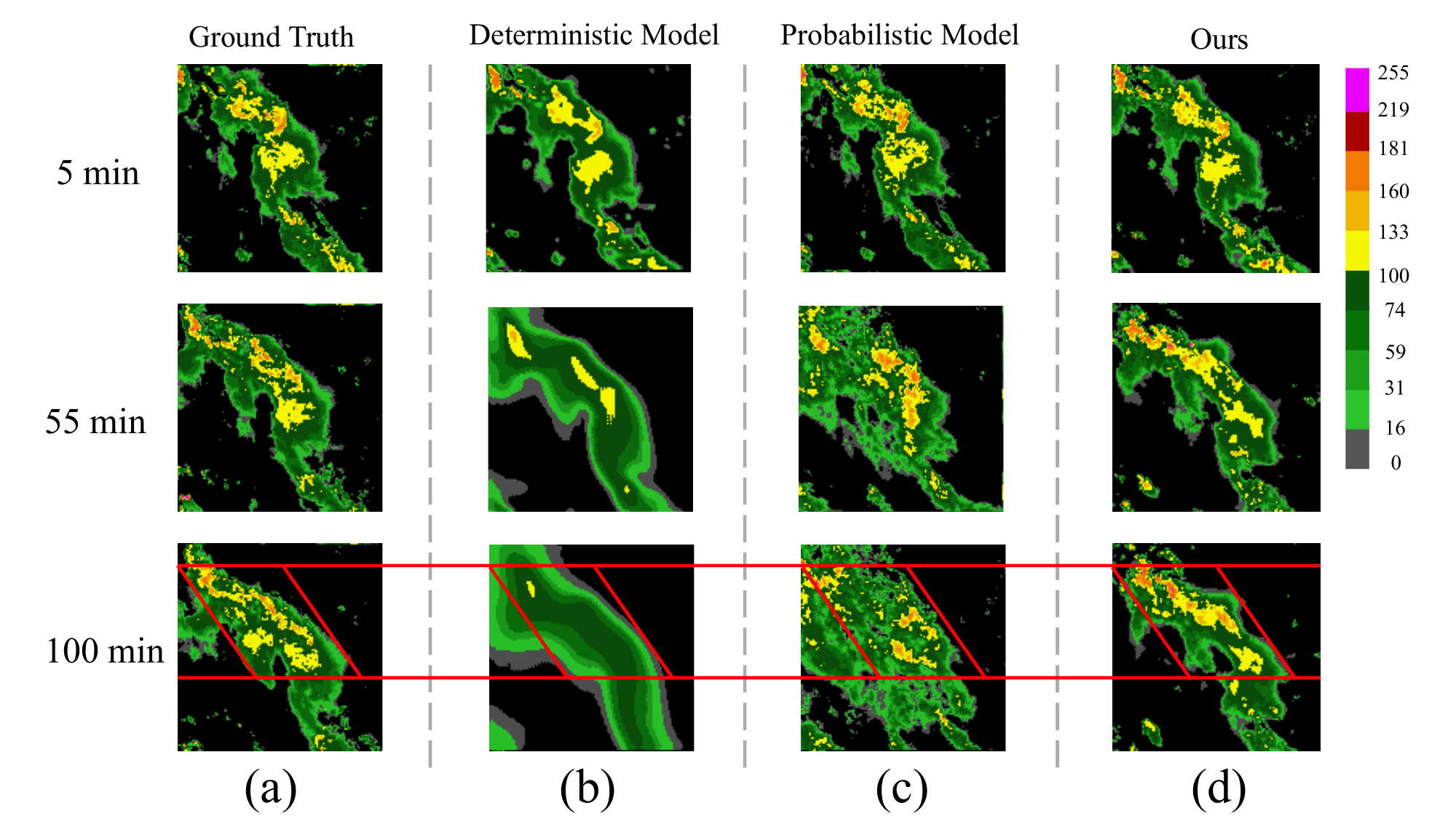}
    \vspace{-4ex}
    
    \caption{We evaluate different models' performances in precipitation nowcasting. The deterministic model (b) can nicely capture the moving trends but tends to get blurry appearance and underestimate the high-value echoes.
    The probabilistic model (c) can capture the appearance details well, but the predicted rainfall positions are inaccurate.
    Our method (d) is able to generate accurate prediction with nice appearance details.
    }
    \label{fig:intro}
    \vspace{-4ex}
\end{figure}

Precipitation nowcasting, aiming at providing a high spatio-temporal resolution rainfall prediction of very short-range (\eg, 0$\sim$6 hours), is a crucial task in meteorological science and smart city applications~\cite{shi2015convolutional,veillette2020sevir}. The core of the task is to predict the future radar echoes sequence based on current observations. In essence, it is a spatio-temporal prediction problem, but with some quite distinguishing traits.

Unlike conventional video or spatio-temporal prediction tasks~\cite{gao2022simvp, lee2018stochastic}, which aim to forecast a deterministic trajectory of moving objects, precipitation particles constitute a chaotic evolutionary system, governed not only by the deterministic moving trends of global system, but also by the stochastic variations (growth or decay) of local particles. 
More importantly, the global moving trends and local variations are inherently coupled. 
For example, when cold and warm air masses cross in a region, continuous rainfall is caused by an evolutionary convection with lasting growth and decay of precipitation particles, rather than a simple movement of seen rainfall belt.

However, most of previous spatio-temporal methods \cite{shi2015convolutional,gao2022simvp,gao2022earthformer, chang2021mau, guen2020disentangling,voleti2022mcvd,gao2023prediff} are developed for conventional tasks, which fail to analyze and model this nature. 
As a result, their performance is unsatisfactory for precipitation nowcasting.
In terms of methodology, we categorize these methods into deterministic and probabilistic models. 
Each type has its own disadvatanges. 
Deterministic models \cite{shi2017deep,gao2022earthformer, gao2022simvp} leverage convolution recurrent neural network, transformer or their variants to forecast the future echoes sequence by directly optimizing the distance or similarity to ground-truth. 
This line of methods can nicely capture the moving trends, but the forecast is getting more and more blurry as the lead time increases, shown as in Figure~\ref{fig:intro}(b). 
Moreover, the high-value echoes, which signify heavy storms, tend to be badly underestimated. 
The reason is that these methods overlook modelling the local stochastics.
To address the blurry issue, researchers recently start to turn to probabilistic generation methods with generative adversarial networks (GANs) \cite{chang2022strpm,zhang2023skilful} or diffusion models \cite{hatanaka2023diffusion,gao2023prediff}. Although the forecast images of such methods are with realistic details, the predicted rainfall positions are often very inaccurate, shown as in Figure~\ref{fig:intro}(c).
This is because these methods model the whole precipitation system in a stochastic manner, where the freedom of generation is too high to maintain a good prediction accuracy.

In this paper, we propose a flexible and unified end-to-end framework, called Diffcast, which nicely decomposes and models the global determinism and local stochastics of precipitation systems.
In the framework, we decompose the evolving system as a motion trend and its stochastic residual, and design a motion component and a temporal residual diffusion component to model them, respectively. 
The motion component serves as a deterministic predictive backbone for global trajectory of precipitation system, and the carefully-designed temporal residual diffusion component accounts for modeling the residual distribution which signifies local stochastics. 
Specially, in the diffusion component, we design a Global Temporal UNet(GTUnet), which carefully utilizes multi-scale temporal features, including the global motion prior, sequence segment consistency and inter frame dependency, as diffusion conditions to model the temporal evolution of residual distribution. 
In the framework, the deterministic predictive component and local stochastic diffusion component are simultaneously trained in an end-to-end manner. 
Hence, they interplay with each other natrually. 
Our framework has three striking advantages. 
First, it is a robust and easy-to-train method without mode collapse compared to adversarial networks. 
Second, any type of deterministic spatio-temporal prediction models (reccurent-based or recurrent-free) can be easily equipped into the framework as predictive backbones. 
Third, the prediction performance of backbones, in terms of both accuracy metrics and appearance details, are significantly improved with our framework.

In summary, our main contributions are summarized as:
\begin{itemize}[leftmargin=0.5cm]
    \item We firstly propose to model the precipitation evolution from the perspective of global deterministic motion and local stochastic variations with residual mechanism. 
    \item We propose a flexible precipitation nowcasting framework that can equip any deterministic backbones (\ie recurrent-based and recurrent-free models) to generate accurate and realistic prediction.
    \item \Change{Inspired by the natural meteorological mechanism, we propose to simultaneously train the deterministic predictive component and the stochastic diffusion component in an end-to-end manner and validate the effectiveness and necessity.}
    \item We equip several notable backbones into our framework and conduct extensive experiments on four publicly available radar echo datasets, and the results show that our framework significantly improves the performance, delivering state-of-the-art results for precipitation nowcasting.
\end{itemize}

%% file: sec/2_formatting.tex
\section{Related Work}

\subsection{Spatio-temporal Predictive Models}

\textbf{Deterministic predictive models} are the mainstream of the existing approaches for spatio-temporal prediction~\cite{ning2023mimo,wu2021motionrnn,bai2022rainformer}. They can be roughly categorized into two groups: recurrent-based models and recurrent-free models. 
The recurrent-based models learn a hidden state from the historical sequence and generate the future frames recurrently with the hidden state \cite{shi2015convolutional,shi2017deep,chang2021mau,wang2017predrnn,tan2023temporal}. 
For example, Shi \etal proposed ConvLSTM \cite{shi2015convolutional} and ConvGRU~\cite{shi2017deep} by integrating the convolution operator into the recurrent neural network. 
MAU \cite{chang2021mau} enhanced the video prediction by developing an attention-based predictive unit with a better temporal receptive field. 
PhyDnet~\cite{guen2020disentangling} incorporated partial differential equation (PDE) constraints in the recurrent hidden state.
The recurrent-free based models~\cite{ning2023mimo,zhang2023skilful} encode the given input frames into hidden states and decode all the predictive frames at once, instead of in a recurrent way. For example, SimVP~\cite{gao2022simvp} designed a scheme to encode and decode the information via simple convolutions. Earthformer~\cite{gao2022earthformer} leveraged transformer to build the encoder-decoder for prediction. 
However, all the deterministic predictive methods suffer from the aforementioned blurry issue and high-value echoes fading away issue for precipitation nowcasting, because of the neglect of local stochastics.

\textbf{Probabilistic predictive models} are designed to capture the spatio-temporal uncertainty by estimating the conditional distribution of future state.
These models aim to enhance the realism of predictions based on adversarial training~\cite{luo2022experimental, chang2022strpm, tulyakov2018mocogan,ravuri2021skilful} or variational autoencoders~\cite{zhang2023skilful,babaeizadeh2018stochastic,tulyakov2018mocogan,franceschi2020stochastic}.
However, these methods directly model the whole precipitation system as stochastics and thus introduce uncontrollable randomness that harms to prediction accuracy. Moreover, these methods often suffer from mode collapse issue and are not easy to train.

\subsection{Spatio-temporal Diffusion Models}
Thanks to the high-fidelity generative capabilities and stable training merit, diffusion models, garnered growing interest for spatio-temporal prediction lately~\cite{jiang2023motiondiffuser,harvey2022flexible,yu2023video,hoppe2022diffusion,mei2023vidm}. 
The mainstream of the methods is based on denoising diffusion probabilistic models~\cite{ho2020denoising,dhariwal2021diffusion}.
For instance, 
MCVD~\cite{voleti2022mcvd} introduced a random mask scheme on frames and utilized conditional denoising network to learn temporal dependency between frames for prediction.
LDCast~\cite{leinonen2023latent} applied a channel-based conditional latent diffusion model to 
predict the evolution of precipitation with denoising mechanism.
PreDiff~\cite{gao2023prediff} incorporated prior knowledge via extra knowledge control network to limit 
the generative process so as to align the prediction with domain-specific prior knowledge based on latent diffusion models. 
However, all the methods regard the whole system evolving stochastically and model it with a diffusion process. In this way, too much freedom is introduced for diffusion models to precisely control the generation. 
As a result, the predictive frames are with realistic details, but the positions tend to be mismatched.



\begin{figure*}[htb]
    \centering
    \includegraphics[width = 2\columnwidth, trim=0 15 5 0, clip]{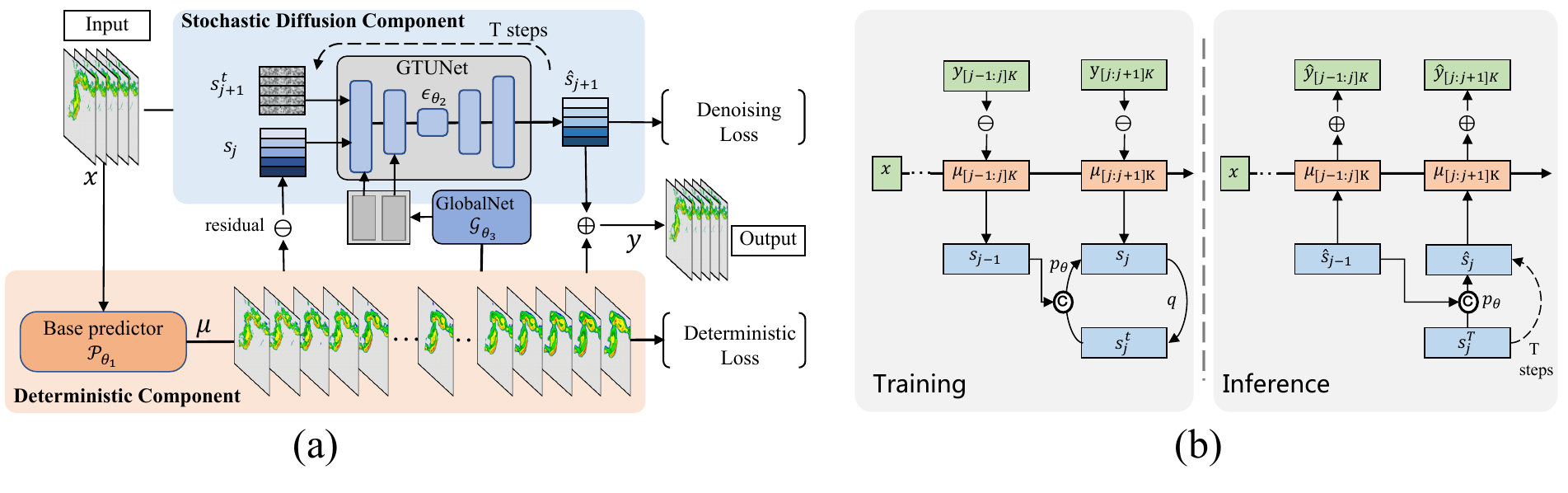}
    \vspace{-2ex}
    
    \caption{The overview of our \textit{DiffCast} framework for precipitation nowcasting is shown in (a). The DiffCast models the precipitation process from two perspective: deterministic component and stochastic component. The former accounts for predicting a global motion trend by a coarse forecast, while the latter aims to incorporate stochasticity with auxiliary-conditioned diffusion into the coarse forecast by residual. \Change{The sub-figure (b)} indicates the computing flow of our framework for training and inference, respectively. The green, orange and blue rectangles represent, respectively, radar echos segment, output of deterministic predictor and residual segment for diffusion model. }
    \label{fig:framework}
    \vspace{-3.5ex}
\end{figure*}

\section{Task Definition and Preliminaries}

\subsection{Task Definition}
We follow ~\cite{veillette2020sevir, gao2023prediff, gao2022earthformer} to formulate precipitation nowcasting as a spatio-temporal prediction problem. Given $L_{in}$ initial frames $ x= [x_i]_{i=-L_{in}}^0 \in \mathbb{R}^{L_{in} \times H \times W \times C}$, the prediction aims to model the conditional probabilistic distribution $p(y|x)$ of the following $L_{out}$ frames $y=[y_i]_{i=1}^{L_{out}} \in \mathbb{R}^{L_{out} \times H \times W \times C}$, where $H$ and $W$ donate the spatial resolution of frame, and $C$ donates the number of measurements at each space-time coordinate. 
In our radar echoes setting, $C=1$ is used.
Note that we denote the $i$-th frame, $y_i$, with a subscript $i$, and use a superscript $t$ to indicate the $t$-th diffusion step state $y^t$ in the following.

\subsection{Preliminary:Diffusion}
Denoising diffusion probabilistic models (DDPMs)~\cite{sohl2015deep, ho2020denoising} 
aim to learn the data distribution $p(y)$ by training a model to reverse a Markov noising process that progressively corrupts the data. 

Specifically, the diffusion model consists of two processes: a forward nosing process and a reverse denoising process. The former is a parameter-free process with progressively noising. We can get the $t$-th diffusion state from the observation $y^0$ with a special property:
\begin{equation}
    \label{eq:q0}
    q(y^t|y^{0}) = \mathcal{N} (y^t;\sqrt{\overline{\alpha}_t}y^{0}, (1-\overline{\alpha}_t)I),
\end{equation}
where $y^T \sim \mathcal{N}(0,1)$ denotes a sample from a pure Gaussian distribution and $y^0$ is a sample from target distribution. The coefficients are defined as $\overline{\alpha}_t=\prod_{t=1}^T \alpha_t,\alpha_t = 1-\beta_t$, where $\beta_t \in (0,1)$ is predefined by an incremental variance schedule. 
As for the reverse denoising process, it is defined by the following Markov chain with parameterized Gaussian transitions:
\begin{equation}
    \label{eq:pt}
    p_{\theta}(y^{t-1}|y^t) = \mathcal{N}(y^{t-1};\mu_{\theta}(y^t, t),\sigma_t^2 I),
\end{equation}
where $\mu_{\theta}$ donates the posterior mean function. 
The reverse function $ p_{\theta}(y^{t-1}|y^t)$ aims to remove the noise added in the forward noising process and we follow~\cite{ho2020denoising} to set the parameterization:
\begin{equation}
    \label{eq:mu_par}
    \mu_{\theta}(y^t,t) = \frac{1}{\sqrt{\alpha_t}}(y^t-\frac{\beta_t}{\sqrt{1-\overline{\alpha}_t}}\epsilon_{\theta}(y^t,t)),
\end{equation}
where $\epsilon_{\theta}(y^t,t)$ is a trainable denoising function that estimates the corresponding noise conditioned on the current step $t$.
The parameters $\theta$ can be optimized by the following loss:
\begin{equation}
    \label{eq:diff_loss}
    L(\theta) = \mathbb{E}_{y^0,t,\epsilon}\| \epsilon-\epsilon_{\theta}(y^t, t)\|^2, 
\end{equation}
where $y^t = \sqrt{\overline{\alpha}_t} y^0 + \sqrt{1-\overline{\alpha}_t} \epsilon$. 
After training, we can recover $y^{t-1}$ from $y^t$ with 
\begin{equation}
    \label{eq:yt-1}
    y^{t-1} = \frac{1}{\sqrt{\alpha_t}}(y^t-\frac{\beta_t}{\sqrt{1-\overline{\alpha_t}}} \epsilon_{\theta}(y^t, t))+\sigma_t \epsilon,
\end{equation}
where $\sigma_t$ is a variance hyperparameter.
Finally, by drawing a sample from the prior distribution $p(y^T)$ and iteratively applying Equation~(\ref{eq:yt-1}), a sample from the target distribution $p(y^0)$ can be derived.



\section{Overall Framework}
Our framework, DiffCast, is designed to decompose the precipitation systems as the global motion trend and local stochastic residual, and utilizes a deterministic component and stochastic diffusion component to model them, respectively. 
The overview of our framework is illustrated in Figure~\ref{fig:framework} (a).  
Specifically, we apply a deterministic backbone as base predictor $\mathcal{P}_{\theta_1}$ to capture the global motion trajectory denoted as $\mu$. Note that any type of deterministic prediction models, \eg SimVP~\cite{rasul2021autoregressive}, Earthformer~\cite{gao2022earthformer}, ConvGRU~\cite{shi2017deep} etc., can be equipped into our framework without additional adaptation (Section \ref{sec:first_stage}).
Then, the deterministic motion prior $\mu$ is leveraged for producing a residual sequence $r$ with ground truth $y$ to represent the local stochastics.
We propose an auxiliary stochastic diffusion component to model the evolution of residual distribution (Section \ref{sec:second_stage}).
In the component, a sophisticated Global Temporal UNet(GTUnet) is designed to exploit the global motion prior, sequence segment consistency and inter-frame dependency for the evolution modelling (Section \ref{sec:unet_block}).
Finally, the probabilistic predictive residual $\hat{r}$ and the deterministic prediction $\mu$ are added to form the final prediction $\hat{y}$. 

\subsection{Deterministic Predictive Backbone}
\label{sec:first_stage}

As a generic module, the deterministic component allows any  prediction models of recurrent-based and recurrent-free architectures to be the base predictor $\mathcal{P}_{\theta_{1}}(\cdot)$.
We denote the output of the base predictor as $\mu = [\mu_i]_{i=1}^{L_{out}} \in \mathbb{R}^{L_{out} \times H \times W \times C}$. 
The deterministic predictive backbone learns to model $p_{\theta_{1}}(\mu|x)$ by using the given input frames $x$ with any pixel-wise lose, \eg  mean-squared error (MSE):
\begin{equation}
    \label{eq:loss_mu}
    \mathcal{L}_{\mathcal{P}} = \mathbb{E}[||\mu - y||^2],
\end{equation}
or other natively designed loss function for the backbone. Here $\theta_1$ means the parameters of this part. We term the loss as deterministic loss in our framework. Though the prediction $\mu$ in this way still suffers from the blurry and high-value echoes fading away issues, it is able to capture the global motion trend, which will help us to decompose the local stochastic residual from the precipitation systems next. Also, $\mu$ will provide necessary information for diffusion component to model the evolution of residual later in Section~\ref{sec:unet_block}.


\subsection{Stochastic Residual Prediction}
\label{sec:second_stage}
As aforementioned, precipitation systems evolve with a global motion trend and local stochastics. The predictive backbone has offered us a global motion trend with $\mu$. Next, we introduce how to model the local stochastics. Our notion is to compute the residual $r$ between the ground-truth $y$ and $\mu$ to represent the local stochastics:
\begin{equation}
    r = y - \mu.
    \label{eq:residual}
\end{equation}
However, the computation of $r$ here involves the ground-truth, which thus cannot be directly used in the prediction. Our idea is using $r$ as supervised information to train a diffusion model to predict the residual evolution. Since $r$ denotes a sequence from 1 to $L_{out}$, we model its evolution in an autoregressive manner. In other words, the diffusion model needs to establish the following distribution:
\begin{equation}
    \label{eq:ri}
    p_{\theta_{2}}(r_i|\hat{r}_{i-1}),
\end{equation}
where $\hat{r}_{i-1}$ indicates the predictive residual of the $(i-1)$-th frame, and $\theta_2$ denotes the parameters of the diffusion model.
Suppose that we use a $T$-step denoising diffusion to model the distribution, following Equation~(\ref{eq:pt}) we have:
\begin{equation}
    \Change{p_{\theta_{2}}(r_i^{0:T}|\hat{r}_{i-1})} = p(r^T)\prod_{t=1}^{T}p_{\theta_2}(r_i^{t-1}|r_i^t, \hat{r}_{i-1}),
\end{equation}
where $r^T \sim \mathcal{N}(0,I)$ and $t$ is the denoising step; and $r_i^{t}$ represents the $t$-th denoising state for the $i$-th frame residual.
Similar to Equation~(\ref{eq:mu_par}), in the denoising process, learning to recover the residual state $r_i^{t-1}$ from state $r_i^{t}$ is equivalent to estimating the corresponding noise $\epsilon$ added in the $t$-th step corruption. Hence, we can use noise estimation as optimization objective.
Suppose that each step of the diffusion process shares the same denoiser function $\epsilon_{\theta_2}(r_{i}^t,\hat{r}_{i-1}, t)$, which takes the previous diffusion state $r^t_{i}$ and the most recent predictive residual $\hat{r}_{i-1}$ as input.
We thus have the following objective function:
\begin{equation}
\label{eq:l2}
    \mathcal{L}_{\epsilon} = \mathbb{E}_{(r_i,r_{i-1})\sim r,t,\epsilon\sim \mathcal{N}(0,I)}||\epsilon-\epsilon_{\theta_2}(r_{i}^t,\hat{r}_{i-1},t)||^2,
\end{equation}
where the $t$-th denoising state $r_i^t$ can be computed as:
\begin{equation}
    r_{i}^t = \sqrt{\overline{\alpha}_t} (\underbrace{y-\mathcal{P}_{\theta_1}(x)}_{residual})_{i} + \sqrt{1-\overline{\alpha}_t}\epsilon.
\end{equation}
We term the objective in Equation~(\ref{eq:l2}) as denoising loss. To capture the interplay between the deterministic predictive backbone and the stochastic residual prediction, we train our framework in an end-to-end manner with the following combined loss function: 
\begin{equation}
    \mathcal{L} = \alpha \sum_{r_{i} \in r}\mathcal{L}_\epsilon(r_{i}) + (1-\alpha) \mathcal{L}_{\mathcal{P}},
    \label{eq:loss_all}
\end{equation}
where $\alpha \in [0,1]$ is a weight factor to balance two losses. 
Once well trained, we can predict the residual $r_i$ by the previous residual $\hat{r}_{i-1}$ with iteratively denoising from a Gaussian noise sample by Equation~(\ref{eq:yt-1}). Repeating this $L_{out}$ times leads us to an estimated residual sequence $\hat{r} = [\hat{ r}_i]^{L_{out}}_{i=1}$.
Note that $\hat{r}_0=0$ in the setting. 
Finally, we compute the final prediction result $\hat{y}$ as: 
 \begin{equation}
     \hat{y} = \hat{r} + \mu.
     \label{eq:hat_y}
 \end{equation}

\subsection{Global Temporal UNet (GTUNet)}
\label{sec:unet_block}
\begin{figure}
    \centering
    \includegraphics[width = 1\columnwidth]{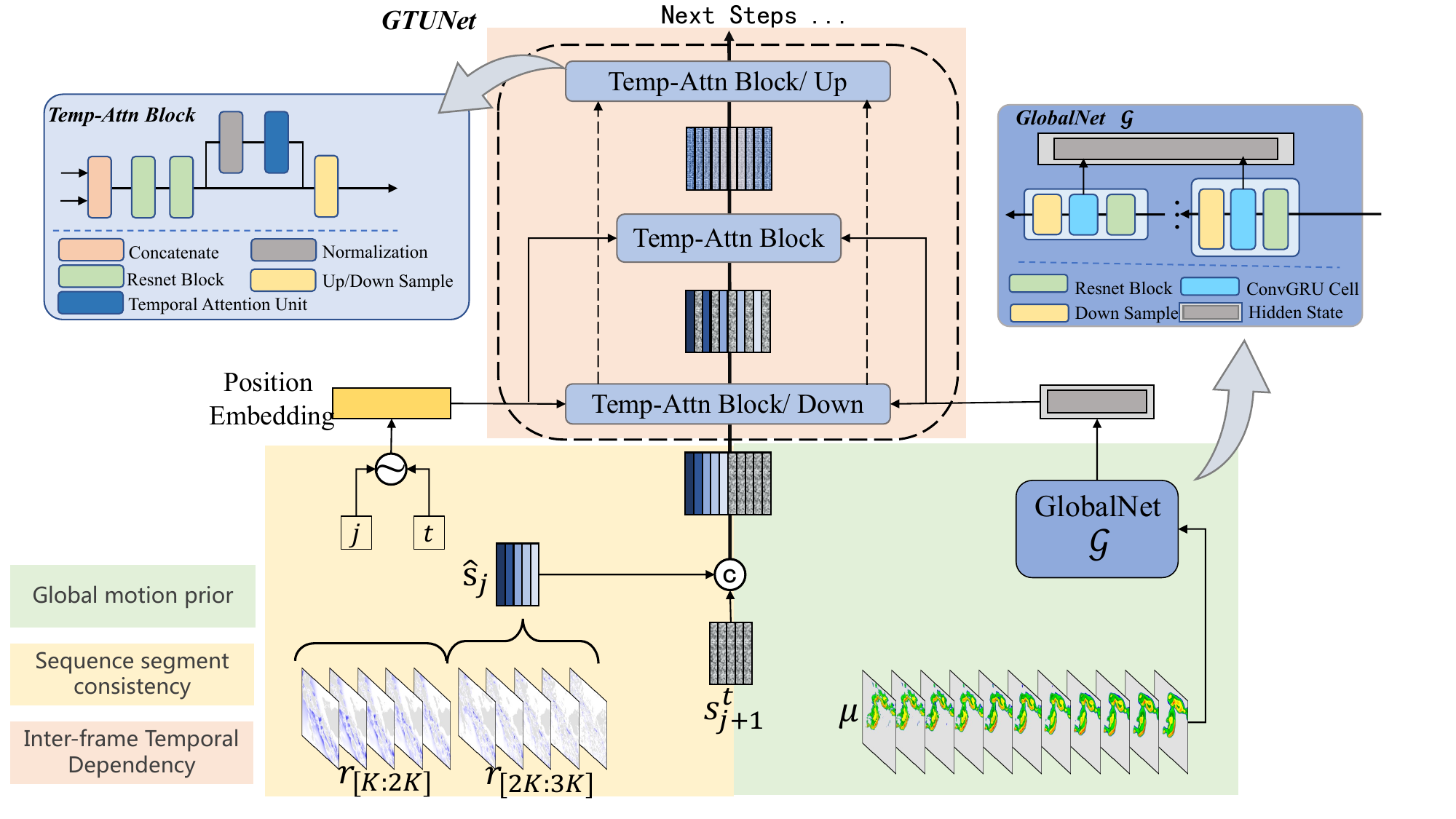}
    \vspace{-4ex}

    \caption{Illustration of our Global Temporal Unet (\ie, GTUnet).}
    \label{fig:unet}
    \vspace{-3.5ex}
\end{figure}

In this subsection, we introduce the detailed diffusion component for the stochastic residual prediction. In the diffusion component we design a Global Temporal UNet, which effectively exploits multi-scale temporal features, namely global motion prior, sequence segment consistency, and inter-frame temporal for the residual evolution prediction. Figure~\ref{fig:unet} illustrates the details of the diffusion model blocks. Next, we elaborate the key blocks, respectively.


\textbf{Global motion prior} refers to the global motion trend information derived from the base deterministic prediction $\mu$. 
As shown in green part of Figure~\ref{fig:unet}, we design a ConvRNN-liked structure, \textbf{GlobalNet} $\mathcal{G}_{\theta_3}$ to extract the global motion information from the deterministic prediction as: 
\begin{equation}
    \label{eq:hs}
    h = \mathcal{G}_{\theta_3}(\mathcal{P}_{\theta_1}(x)).
\end{equation}
Here $\theta_3$ represents the parameters of the \textbf{GlobalNet}. As shown in dark blue part of Figure~\ref{fig:unet}, \textbf{GlobalNet} is a multi-layer architecture with multiple temporal blocks. Each temporal block is composed of a down sample operator, a ConvGRU operator and a Resnet operator. By incorporating the derived hidden state $h$ as an extra condition into diffusion model, the residual prediction $r$ is re-expressed as:
\begin{equation}
    r = [p_{\theta_2}(r_{i}|\hat{r}_{i-1},h)]_{i=1}^{L_{out}},
\end{equation}
and the denoiser function is changed into:
\begin{equation}
\label{eq:denoising}
    \epsilon_{\theta_2}(r_{i}^t, \hat{r}_{i-1}, \mathcal{G}_{\theta_3}(\mathcal{P}_{\theta_1}(x)),t).
\end{equation}
The objective function in Equation (\ref{eq:loss_all}) is then parameterized by $(\theta_1, \theta_2, \theta_3)$.



\textbf{Sequence segment consistency}. To better maintain the sequence consistency of residual evolution, we propose to partition the residual sequence $r$ into multiple segments $s$ for prediction, because recent studies~\cite{ning2023mimo} show that multi-input multi-output is a better paradigm than single-in single-out in recurrent spatio-temporal prediction. 
As shown in yellow part of Figure~\ref{fig:unet}, we construct the segment and denote the $j$-th segment as:
\begin{equation}
    s_{j-1} = r_{[(j-1)K:jK]},
    \label{eq:s_j-1}
\end{equation}
where $K$ indicates the length of each segment and $j = 0, 1,...,\lceil\frac{L_\text{out}}{K}\rceil$.
Then we have $s_j \in \mathbb{R}^{K \times H \times W \times C}$.
In this way, the diffusion is changed to model the segment-level temporal distribution as:
\begin{equation}
    s = [p_{\theta_2}(s_j|\hat{s}_{j-1},h)]_{j=1}^{\lceil\frac{L_\text{out}}{K}\rceil}.
\end{equation}
Specifically, we incorporate the segment condition with channel concatenation between $t$-th denoising state $s_{j}^{t}$ and previous segment $s_{j-1}$.
Moreover, as the forecast lead time increases, the residual inevitably becomes larger because of the increased uncertainty. 
Hence, it is important to explicitly indicate the position of a segment. 
To this end, we add an extra position embedding on segment index $j$, in addition to the denoising step $t$.
With the above changes, the objective of diffusion model for residual evolution becomes:
\begin{equation}
    \label{eq:l_ep}
        \mathcal{L}_{\epsilon} = \mathbb{E}||\epsilon-\epsilon_{\theta_2}(s_{j}^t,\hat{s}_{j-1},\mathcal{G}_{\theta_3}(\mathcal{P}_{\theta_1}(x)),t,j)||^2.
\end{equation}
Note that when predicting the first residual segment $s_1$, we use $s_0$=0 following~\cite{voleti2022mcvd}.


\textbf{Inter-frame temporal dependency}. 
To better model the inter-frame  dependency within a segment, our Global Temporal UNet is carefully designed with temporal attention blocks, as shown in orange part of Figure~\ref{fig:unet}. It is indeed a variant of UNet in DDPM with temporal evolution. The \textbf{Temp-Attn Block}, as shown in grey blue part of Figure~\ref{fig:unet}, is constructed by concatenate operator, Resnet operator, normalization operator and temporal attention unit operator, followed by an up/downsample operator, which makes the predicted residual in the segment become temporal dependent. With the developed temporal evolution UNet structure, the inter-frame  dependency can be effectively taken into account.

\subsection{Training and Inference}
For clarity, we summarize the computing flow of training and inference of our framework in Figure~\ref{fig:framework} (b). In the training phase, the deterministic predictor $\mathcal{P}_{\theta_1}$ first produces $\mu$. Then the residual $r$ is computed as Equation (\ref{eq:residual}) and grouped into segments $s$ by Equation (\ref{eq:s_j-1}). Given a segment $s_{j-1}$, we predict the segment $s_j$ by our diffusion model $p_{\theta_2}$. Here the diffusion model is updated conventionally, namely sampling a step $t$, adding noise as scheduled, using the denoiser to predict noise and calculating loss as Equation (\ref{eq:l_ep}) to update parameters $(\theta_1,\theta_2,\theta_3)$. As for inference, the computing flow is similar with the only difference of diffusion part, where we first sample the $T$-state segment $s_j^{T}$ from $\mathcal{N}(0,I)$ and perform $T$-step denoising with $p_{\theta_2}$ to recover $\hat{s}_{j}$ given $\hat{s}_{j-1}$. Once all required residual segments are obtained we can compute the final prediction as Equation (\ref{eq:hat_y}).

\begin{table*}[ht]
    \caption{Experiment results on four radar datasets. Relative improvements are shown with brackets.}
    \vspace{-2ex}
    \centering
    \resizebox{1.95\columnwidth}{!}{
    \begin{tabular}{c|cccccc|cccccc}
\toprule
\multirow{2}*{Method}&\multicolumn{6}{c|}{\textbf{SEVIR}}& \multicolumn{6}{c}{\textbf{MeteoNet}}\\
\cline{2-13}
& $\uparrow$CSI& $\uparrow$CSI-pool4&  $\uparrow$CSI-pool16&  $\uparrow$HSS&   $\downarrow$LPIPS&  $\uparrow$SSIM&  $\uparrow$CSI&  $\uparrow$CSI-pool4& $\uparrow$CSI-pool16& $\uparrow$HSS&  $\downarrow$LPIPS& $\uparrow$SSIM\\
\midrule
 SimVP\cite{gao2022simvp}& 0.2662 & 0.2844 & 0.3452 & 0.3369  & 0.3914 & 0.6304 
 & 0.3346& 0.3383& 0.4143& 0.4568 & 0.3523& 0.7557
\\
\textbf{DiffCast\_SimVP}& 
\textbf{\makecell{0.3077\\(+15.59\%)}}&
\textbf{\makecell{0.4122\\(+44.94\%)}}&
\textbf{\makecell{0.5683\\(+64.63\%)}}&
\textbf{\makecell{0.4033\\(+19.71\%)}}&
\textbf{\makecell{0.1812\\(+53.70\%)}}&
\textbf{\makecell{0.6354\\(+0.79\%)}}&

\textbf{\makecell{0.3511\\(+4.93\%)}}&
\textbf{\makecell{0.5081\\(+50.19\%)}}&
\textbf{\makecell{0.7155\\(+72.70\%)}}& 
\textbf{\makecell{0.4846\\(+6.09\%)}}& 
\textbf{\makecell{0.1198\\(+65.99\%)}}& 
\textbf{\makecell{0.7887\\(+4.37\%)}}
\\
\hline
 Earthformer\cite{gao2022earthformer}& 0.2513& 0.2617& 0.2910& 0.3073&  0.4140& \textbf{0.6773}&
0.3296& 0.3428& 0.4333& 0.4604& 0.3718& 0.7899
\\
\textbf{DiffCast\_Earthformer}&
\textbf{\makecell{0.2823\\(+12.34\%)}}& 
\textbf{\makecell{0.3868\\(+47.80\%)}}&
\textbf{\makecell{0.5362\\(+84.26\%)}}&
\textbf{\makecell{0.3623\\(+17.90\%)}}&

\textbf{\makecell{0.1818\\(+56.09\%)}}&
\makecell{0.6420\\(-5.21\%)}&
\textbf{\makecell{0.3402\\(+3.22\%)}}&
\textbf{\makecell{0.5020\\(+46.44\%)}}&
\textbf{\makecell{0.7092\\(+63.67\%)}}&
\textbf{\makecell{0.4696\\(+2.00\%)}}&

\textbf{\makecell{0.1236\\(+66.76\%)}}& 
\textbf{\makecell{0.7967\\(+0.86\%)}}
\\
\hline
 MAU\cite{chang2021mau}& 0.2463& 0.2566& 0.2861& 0.3004& 0.3933& 0.6361
 & 0.3232& 0.3304& 0.4165& 0.4451 & 0.3089& 0.7897
\\
\textbf{DiffCast\_MAU}&
\textbf{\makecell{0.2716\\(+10.27\%)}}& 
\textbf{\makecell{0.3789\\(+47.66\%)}}&
\textbf{\makecell{0.5414\\(+89.23\%)}}&
\textbf{\makecell{0.3506\\(+16.71\%)}}&

\textbf{\makecell{0.1874\\(+52.35\%)}}&
\textbf{\makecell{0.6729\\(+5.79\%)}}&
\textbf{\makecell{0.3490\\(+7.98\%)}}&
\textbf{\makecell{0.5030\\(+52.24\%)}}&
\textbf{\makecell{0.7114\\(+70.80\%)}}&
\textbf{\makecell{0.4822\\(+8.34\%)}}& 
\textbf{\makecell{0.1213\\(+60.73\%)}}& 
\makecell{0.7665\\(-2.94\%)}
\\
\hline
 ConvGRU\cite{shi2017deep}& 0.2416& 0.2554& 0.3050& 0.2834&  0.3766& \textbf{0.6532}& 
 0.3400& 0.3578& 0.4473& 0.4667&  0.2950& 
\textbf{0.7832}
\\
\textbf{DiffCast\_ConvGRU}& 
\textbf{\makecell{0.2772\\(+14.74\%)}}&
\textbf{\makecell{0.3809\\(+49.14\%)}}& 
\textbf{\makecell{0.5463\\(+79.11\%)}}&
\textbf{\makecell{0.3551\\(+25.30\%)}}&

\textbf{\makecell{0.1880\\(+50.08\%)}}& 
\makecell{0.6188\\(-5.27\%)}&
\textbf{\makecell{0.3512\\(+3.29\%)}}&
\textbf{\makecell{0.4930\\(+37.79\%)}}&
\textbf{\makecell{0.7001\\(+56.52\%)}}&
\textbf{\makecell{0.4862\\(+4.18\%)}}&

\textbf{\makecell{0.1244\\(+57.83\%)}}&
\makecell{0.7761\\(-0.91\%)}
\\
\hline
 PhyDnet\cite{guen2020disentangling}&
 0.2560&0.2685 &0.3005 &0.3124  &0.3785 &\textbf{0.6764} 
 & 0.3384& 0.3824& 0.4986& 0.4673& 0.2941& \textbf{0.8022}
\\
\textbf{DiffCast\_PhyDnet}&
\textbf{\makecell{0.2757\\(+7.70\%)}}&
\textbf{\makecell{0.3797\\(+41.42\%)}}&
\textbf{\makecell{0.5296\\(+76.24\%)}}&
\textbf{\makecell{0.3584\\(+14.72\%}}&

\textbf{\makecell{0.1845\\(+51.2\%)}}&
\makecell{0.6320\\(-6.56\%)}&

\textbf{\makecell{0.3472\\(+2.60\%)}}&
\textbf{\makecell{0.5066\\(+32.48\%)}}&
\textbf{\makecell{0.7200\\(+44.40\%)}}&
\textbf{\makecell{0.4802\\(+2.76\%)}}&

\textbf{\makecell{0.1234\\(+58.04\%)}}&
\makecell{0.7788\\(-2.92\%)}
\\
\midrule
MCVD\cite{voleti2022mcvd}&0.2148  &0.3020  & 0.4706 & 0.2743 &0.2170  &0.5265  &0.2336  &0.3841  &0.6128 &0.3393  &0.1652 &0.5414 \\
PreDiff\cite{gao2023prediff}&0.2304  &0.3041  &0.4028  &0.2986   &0.2851  &0.5185  
&0.2657  &0.3854  &0.5692 &0.3782&0.1543 &0.7059 \\
STRPM\cite{chang2022strpm}&0.2512  &0.3243  & 0.4959  & 0.3277   &0.2577  &0.6513
&0.2606  & 0.4138  &0.6882 & 0.3688 &0.2004 &0.5996 \\
\bottomrule     
\toprule
 \multirow{2}*{Method}&  \multicolumn{6}{c|}{\textbf{Shanghai\_Radar}}&  \multicolumn{6}{c}{\textbf{CIKM}}\\
 \cline{2-13}
         & $\uparrow$CSI& $\uparrow$CSI-pool4&  $\uparrow$CSI-pool16&  $\uparrow$HSS&   $\downarrow$LPIPS&  $\uparrow$SSIM&  $\uparrow$CSI&  $\uparrow$CSI-pool4& $\uparrow$CSI-pool16& $\uparrow$HSS&  $\downarrow$LPIPS& $\uparrow$SSIM\\
         \midrule
SimVP\cite{gao2022simvp}&  0.3841&  0.4467&  0.5603&  0.5183&  0.2984&  0.7764
&\textbf{0.3021}&0.3530&0.4677&\textbf{0.3948}& 0.3134& 0.6324
\\
\textbf{DiffCast\_SimVP}&  
\textbf{\makecell{0.3955\\(+2.97\%)}}&
\textbf{\makecell{0.5116\\(+14.53\%)}}&  
\textbf{\makecell{0.6576\\(+17.37\%)}}&
\textbf{\makecell{0.5296\\(+2.18\%)}}&

\textbf{\makecell{0.1571\\(+47.35\%)}}& 
\textbf{\makecell{0.7902\\(+1.78\%)}}&

\makecell{0.2999\\(-0.73\%)}&
\textbf{\makecell{0.3657\\(+3.60\%)}}&  
\textbf{\makecell{0.5260\\(+12.47\%)}}&
\makecell{0.3874\\(-1.87\%)}&
\textbf{\makecell{0.2223\\(+29.07\%)}}& 
\textbf{\makecell{0.6391\\(+1.06\%)}}
\\
\hline
Earthformer\cite{gao2022earthformer}&  0.3575&  0.4008&  0.4863&  0.4843&    0.2564&0.7750
&\textbf{0.3153}& 0.3547& 0.4927& 0.3828 & 0.3857& \textbf{0.6510}
\\
\textbf{DiffCast\_Earthformer}&
\textbf{\makecell{0.3751\\(+4.92\%)}}&
\textbf{\makecell{0.4855\\(+21.13\%)}}&
\textbf{\makecell{0.6212\\(+27.74\%)}}&
\textbf{\makecell{0.5069\\(+4.67\%)}}&
\textbf{\makecell{0.1586\\(+38.14\%)}}&
\textbf{\makecell{0.7851\\(+1.30\%)}}&

\makecell{0.3099 \\(-1.71\%)}&
\textbf{\makecell{0.3807\\(+7.33\%)}}&
\textbf{\makecell{0.5509\\(+11.81\%)}}&
\textbf{\makecell{0.3947\\(+3.11\%)}}&
\textbf{\makecell{0.2259\\(+41.43\%)}}&
\makecell{0.6313\\(-3.03\%)}

\\
\hline
MAU\cite{chang2021mau}& 0.3996& 0.4695& 0.5787& 0.5356& 0.2735& 0.7303&
0.2936&0.3152& 0.4144&  0.3660 & 0.3999&0.6277
\\
\textbf{DiffCast\_MAU}&
\textbf{\makecell{0.4089\\(+2.33\%)}}&
\textbf{\makecell{0.5212\\(+11.01\%)}}&
\textbf{\makecell{0.6658\\(+15.05\%)}}&
\textbf{\makecell{0.5475\\(+2.22\%)}}&
\textbf{\makecell{0.1618\\(+40.84\%)}}&
\textbf{\makecell{0.7879\\(+7.89\%)}}&

\textbf{\makecell{0.3158\\(+7.56\%)}}&
\textbf{\makecell{0.3803\\(+20.65\%)}}&
\textbf{\makecell{0.5443\\(+31.35\%)}}&
\textbf{\makecell{0.4085\\(+11.61\%)}}& 
\textbf{\makecell{0.2205\\(+44.86\%)}}&
\textbf{\makecell{0.6498\\(+3.52\%)}}
\\
\hline
 ConvGRU\cite{shi2017deep}& 0.3612& 0.4439& 0.5596& 0.4899& 0.2564& 0.7795
&0.3092 &0.3533 &0.4686 &\textbf{0.4007} &0.3135 &0.6601
\\
\textbf{DiffCast\_ConvGRU}&
\textbf{\makecell{0.3738 \\(+3.49\%)}}&
\textbf{\makecell{0.4923 \\(+10.90\%)}}&
\textbf{\makecell{0.6596\\(+17.87\%)}}&
\textbf{\makecell{0.4945\\(+0.94\%)}}&
\textbf{\makecell{0.1563\\(+39.04\%)}}&
\textbf{\makecell{0.7809\\(+0.18\%)}}&

\textbf{\makecell{0.3143\\(+1.65\%)}}&
\textbf{\makecell{0.3681\\(+4.19\%)}}&
\textbf{\makecell{0.5117\\(+9.20\%)}}&
\makecell{0.3967\\(-1.00\%)}& 

\textbf{\makecell{0.2201\\(+29.79\%)}}&
\textbf{\makecell{0.6418\\(+2.77\%)}}
\\
\hline
PhyDnet\cite{guen2020disentangling}& 0.3653& 0.4552& 0.5980& 0.4957& 0.1894& 0.7751
&0.3037 &0.3442 &0.4655 &0.3931 &0.3631 &\textbf{0.6540}\\
\textbf{DiffCast\_PhyDnet}&
\textbf{\makecell{0.3671\\(+0.49\%)}}&
\textbf{\makecell{0.4907\\(+7.80\%)}}&
\textbf{\makecell{0.6493\\(+8.58\%)}}&
\textbf{\makecell{0.4986\\(+0.59\%)}}&

\textbf{\makecell{0.1574\\(+16.90\%)}}&
\textbf{\makecell{0.7780\\(+0.37\%)}}&
\textbf{\makecell{0.3131\\(+3.10\%)}}&
\textbf{\makecell{0.3836\\(+11.45\%)}}&
\textbf{\makecell{0.5550\\(+19.23\%)}}&
\textbf{\makecell{0.3990\\(+1.50\%)}}&

\textbf{\makecell{0.2270\\(+37.48\%)}}&
\makecell{0.6156\\(-5.87\%)}
\\
\midrule
 MCVD\cite{voleti2022mcvd}&0.2872 &0.3984 &0.5675 &0.4036 &0.2081 &0.5119
 &0.2513 &0.3095 &0.4955 &0.3294 &0.2528 &0.5358\\
 PreDiff\cite{gao2023prediff}&0.3583 &0.4389 &0.5448 &0.4849 &0.1696 &0.7557 
 &0.3043 &0.3681 &0.5117 &0.3967  &0.2201 &0.6418\\
 STRPM\cite{chang2022strpm}&0.3606 &0.4944 &0.6783 &0.4931  &0.1681 &0.7724 &0.2984 &0.3590 &0.5020 &0.3870 &0.2397 & 0.6443\\
    \bottomrule    
    \end{tabular}

}

    \label{tab:all_res}
    \vspace{-3ex}
\end{table*}

\section{Experiments}

\subsection{Experimental Setting}

\textbf{Dataset.}
The \textbf{SEVIR}~\cite{veillette2020sevir}, as a widely used dataset for precipitation nowcasting, contains 20,393 weather events of radar frame sequence with a length of 4-hour and size of 384 km$\times$384 km, where every pixel stands for 1km$\times$1km and the temporal resolution is 5 minutes. 
The \textbf{MeteoNet}~\cite{2020meteo} covers a large area of 550 km$\times$550 km in France, and records over three years observations with temporal resolution of 6 minutes.
The \textbf{Shanghai\_Radar}~\cite{chen2020deep} is generated by volume scans in intervals of approximately 6 minutes from 2015 to 2018 in Pudong, Shanghai with a spatial size of 501 km $\times$501 km. 
The \textbf{CIKM}\footnote{https://tianchi.aliyun.com/dataset/1085} records precipitation events in 101 km$\times$101 km area of Guangdong, China. Each sequence settles 15 radar echo maps as a sample and the temporal resolution is 6 minutes.
More details of these datasets are provided in Appendix.

\begin{figure*}[t]
  \centering
 \begin{subfigure}{0.3\linewidth}
    \includegraphics[width=1\columnwidth]{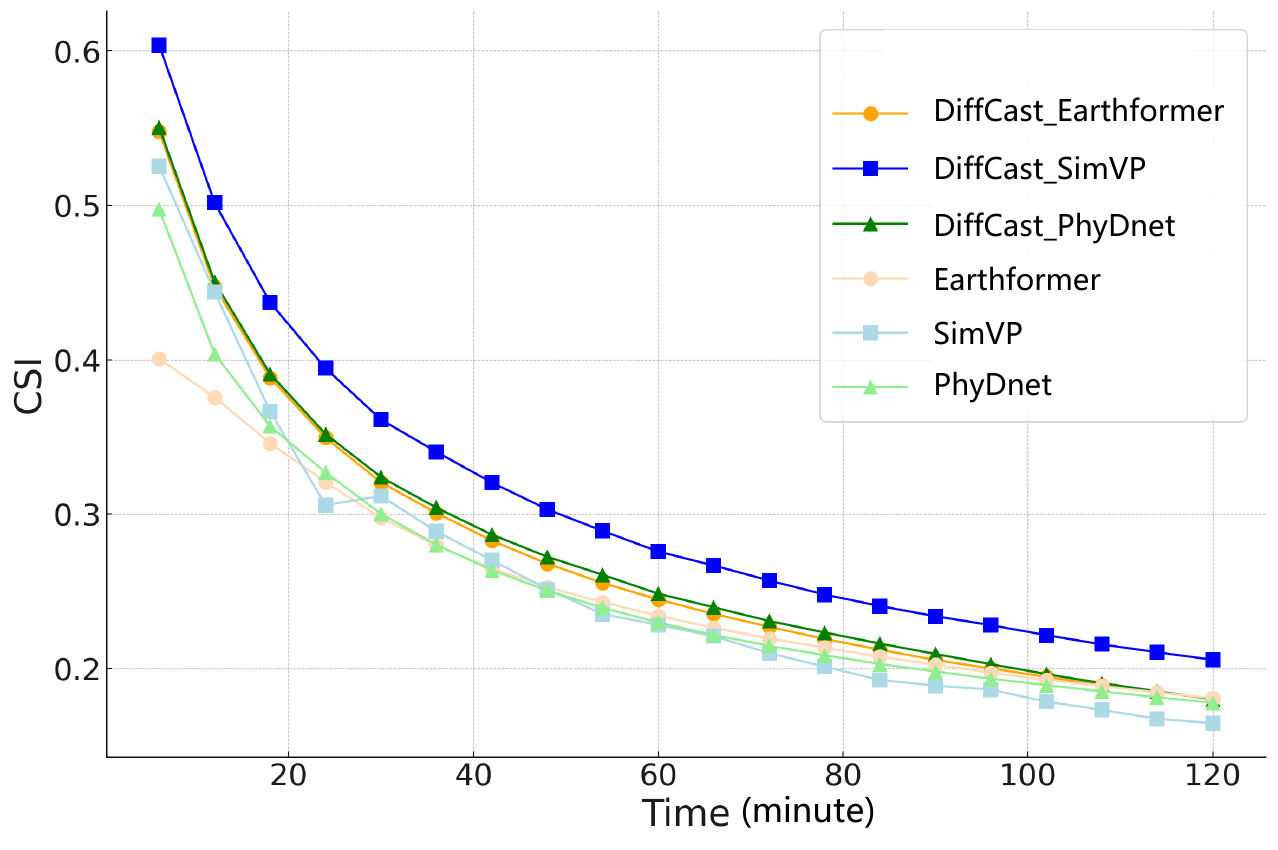}
    \caption{Frame-wise CSI}
    \label{fig:csi}
  \end{subfigure}
  \begin{subfigure}{0.3\linewidth}
    \includegraphics[width=1\columnwidth]{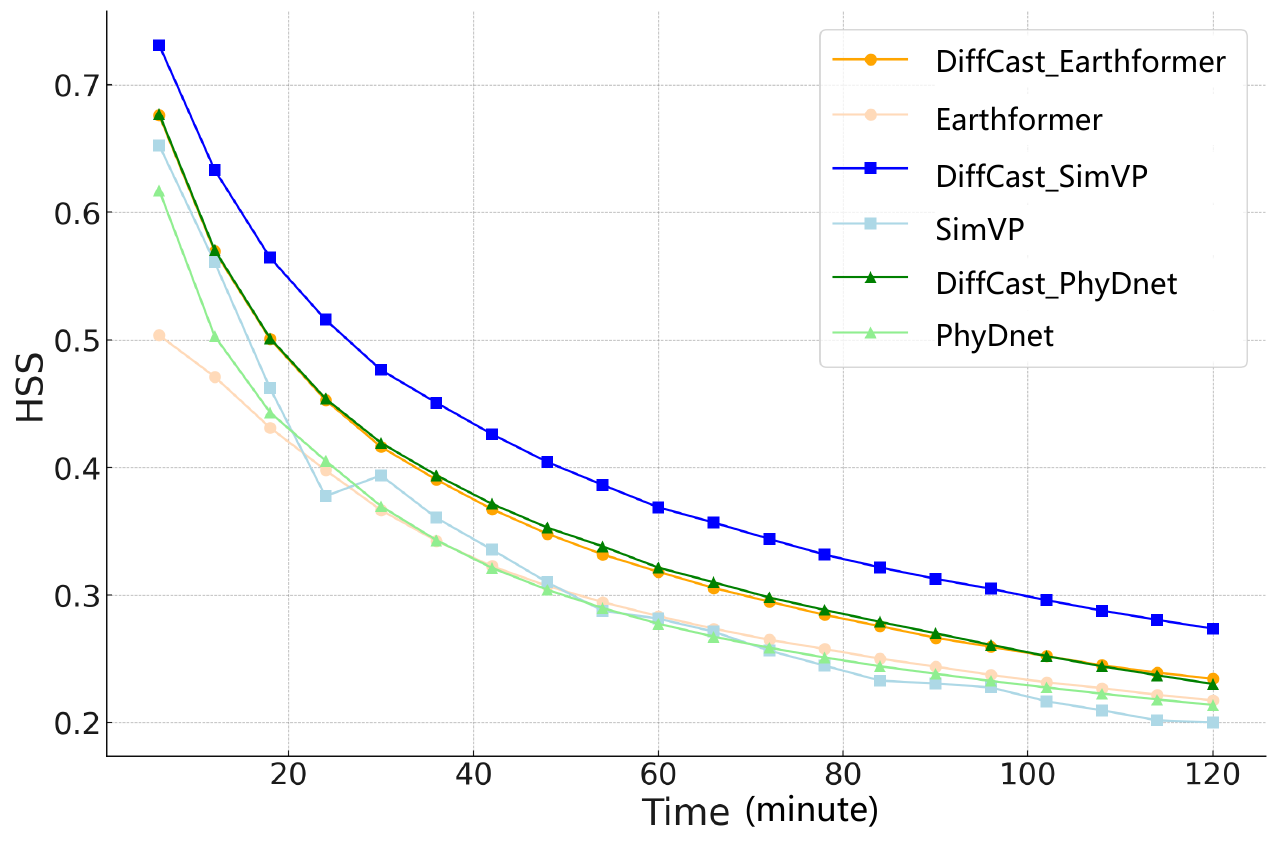}
    \caption{Frame-wise HSS}
    \label{fig:hss}
  \end{subfigure}
  \begin{subfigure}{0.3\linewidth}
    \includegraphics[width=1\columnwidth]{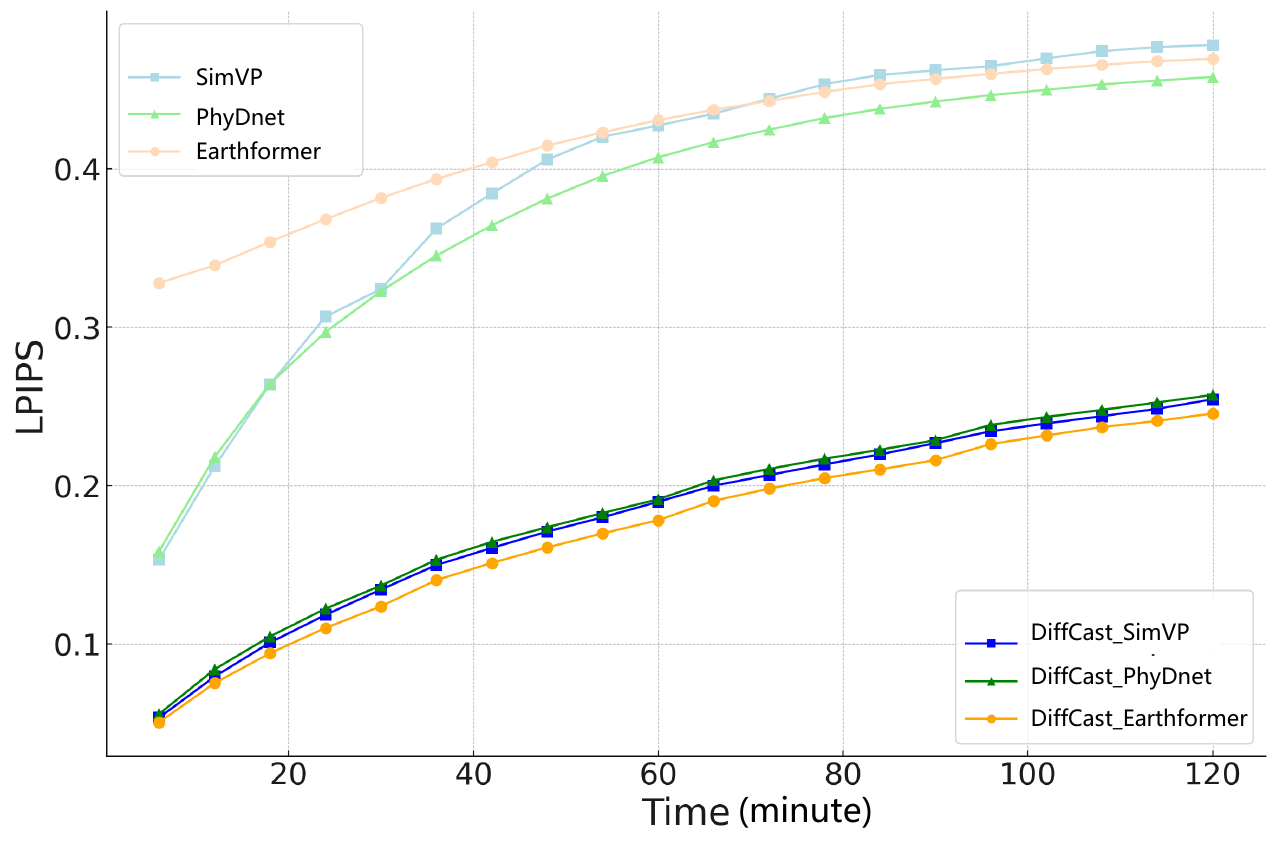}
    \caption{Frame-wise LIPIS}
    \label{fig:lpips}
  \end{subfigure}
  \vspace{-2ex}
  \caption{Performance changes against different lead time in terms of CSI, HSS and LPIPS. For a better vision, we only show the curves of SimVP, Earthformer and PhyDnet with or without our framework.}
  \label{fig:all_res}
  \vspace{-3ex}
\end{figure*}

\textbf{Data preprocess.}
As for all sequences, we mainly focus on modeling the precipitation event. 
Hence, following~\cite{chen2020deep}, we separate the continuous sequence into multiple events for MeteoNet and Shanghai-Radar corpus.
As mentioned in~\cite{zhang2023skilful} that the increasing length of initial frames $L_{in}$ cannot provide substantial improvements for forecast, we set the challenging prediction task to predict 20 frames given 5 initial frames (\ie $5\rightarrow20$) except for the CIKM dataset, where only $5\rightarrow10$ can be used due to its sequence length limitation.
For all the datasets, we keep the original temporal resolution but downscale the spatial size to $128\times128$, due to the limitation of our computation resource.

\textbf{Evaluation.}
To evaluate the accuracy of nowcasting, we calculate the average Critical Success Index (CSI) and Heidke Skill Score (HSS) following~\cite{luo2022reconstitution, gao2023prediff, veillette2020sevir} at different thresholds (the detailed way to compute CSI and HSS can be found at appendix).
The CSI, similar to IoU, is to measure the degree of pixel-wise matching between predictions and ground truth after thresholding them into 0/1 matrices. 
Following~\cite{gao2022earthformer, gao2023prediff}, we also report the CSIs at different pooling scales, which relax the pixel-wise matching to evaluate the accuracy on neighborhood aggregations.
Additionally, LPIPS and SSIM are also utilized to measure the visual quality of prediction.

\textbf{Training details.} We train our DiffCast framework for 200K iterations using Adam optimizer with a learning rate of 0.0001. As for diffusion setting, we follow the standard setting of diffusion model in~\cite{ho2020denoising} to set the diffusion steps as 1000 and denoising steps for inference as 250 with DDIM~\cite{song2020denoising}. We set default loss weight factor $\alpha=0.5$ to balance the deterministic loss and denoisng loss in Equation (\ref{eq:loss_all}).
As for baseline methods, their configurations are tuned correspondingly for different datasets.
All experiments run on a computer with a single A6000 GPU.

\subsection{Experimental Results}

\begin{figure}[t]
    \centering
    \includegraphics[width=1\linewidth, trim=0 20 240 20, clip]{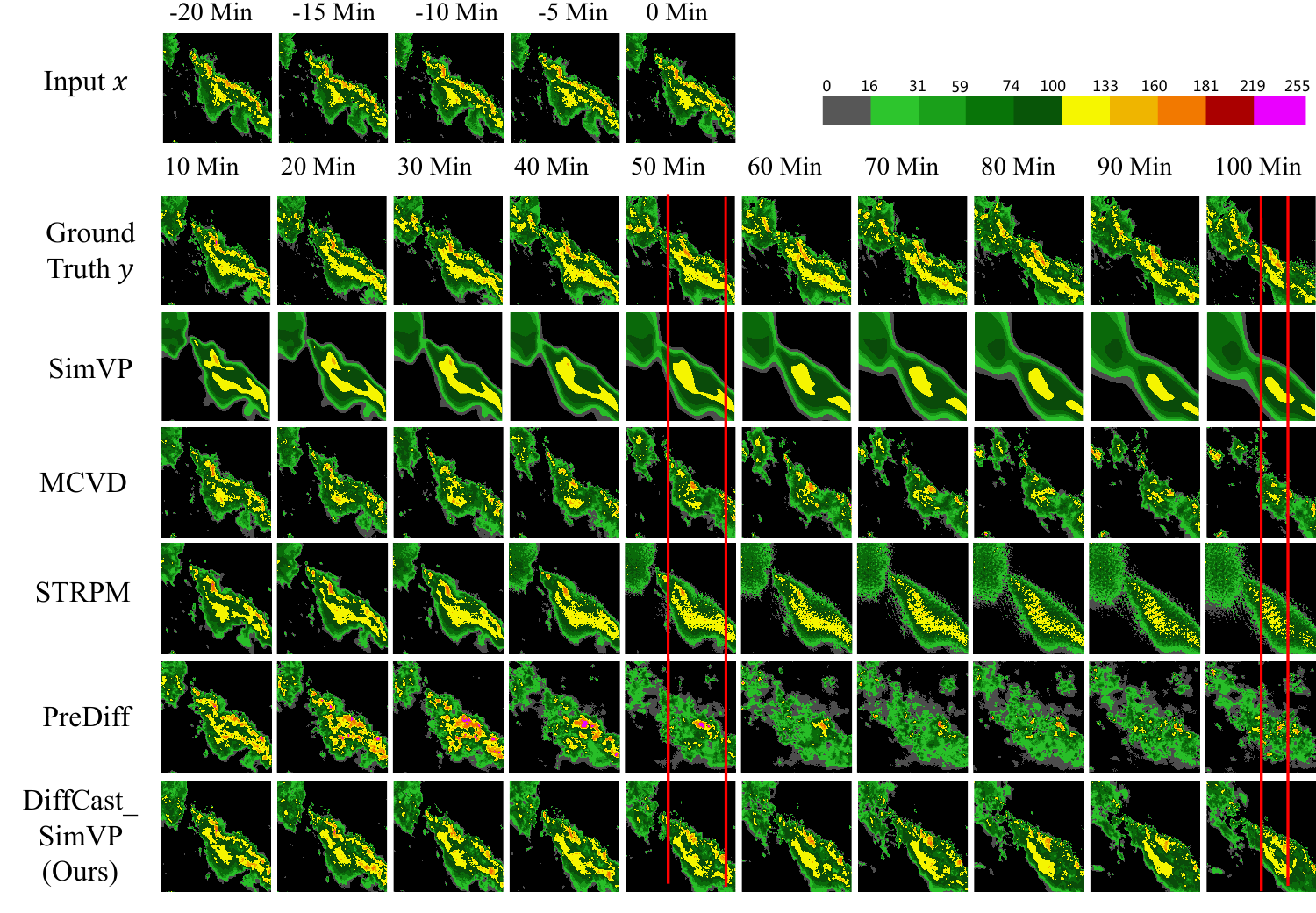}
    \vspace{-4ex}
    \caption{A visual comparison example on a precipitation event from SEVIR. The results of Earthformer, MAE, ConvGRU are similar, which is put into Appendix due to the space limitation.}
    \label{fig:sample1}
    \vspace{-2ex}
\end{figure}
\begin{table}[t]
    \centering
    \caption{Analysis of the necessarity of deterministic loss.}
    \vspace{-2ex}
    \resizebox{0.95\columnwidth}{!}{
    \begin{tabular}{ccccccc}
    \toprule
         Method&  CSI&  CSI-pool4&  CSI-pool16&  HSS&    LPIPS& SSIM\\
          \midrule
 DiffCast\_SimVP($\alpha$=0)& 0.2578& 0.2726& 0.3049& 0.3154&  0.3849&0.6570\\

         DiffCast\_Simvp($\alpha$=0.5)&  \textbf{0.3077}&  \textbf{0.4122}&  \textbf{0.5683}&  \textbf{0.4033}&    \textbf{0.1812}& 0.6354\\
         DiffCast\_Simvp($\alpha$=1)&  0.2430&  0.3065&  0.3999&  0.2989&   0.1831& \textbf{0.6824}\\
    \bottomrule
    \end{tabular}
    }
    \label{tab:d_loss}
    \vspace{-3ex}
\end{table}

As mentioned in Section~\ref{sec:first_stage} that our DiffCast can flexibly utilize various type of deterministic models as a base predictor, we select some notable deterministic models as our comparison baselines. Among them, the SimVP~\cite{gao2022simvp} and Earthformer~\cite{gao2022earthformer} apply recurrent-free strategy to generate all frames at once, while the MAU~\cite{chang2021mau}, ConvGRU~\cite{shi2017deep} and PhyDnet~\cite{guen2020disentangling} are designed with recurrent strategy to generate frames one by one. Moreover, we also utilize two diffusion-based approaches~\cite{voleti2022mcvd, gao2023prediff} and a GAN-based model~\cite{chang2022strpm} for comparison. 
We evaluate all this deterministic models and stochastic generative models, as well as our framework equipped with every deterministic model on four real-world precipitation datasets.  
The experimental results are shown in Table~\ref{tab:all_res}.

From the results of Table~\ref{tab:all_res}, we make the following observations: (i) Equipped into our DiffCast framework, the performances of backbones are significantly improved and the improvements are from $2\%$ to 20$\%$, in terms of CSI and HSS, and more improvements can be observed for pooling CSI. This verifies the effectiveness of our framework to boost the prediction accuracy of backbones. (ii) In terms of LPIPS and SSIM, which measures the visual quality of predictions, our DiffCast framework also makes a significant improvement. Especially for LPIPS,  the  improvements are $16.9\%\sim 66.7\%$. This suggests that our framework indeed improves the visual quality of backbone predictors. (iii) Compared to the state-of-the-art GAN and diffusion baselines, namely MCVD, PreDiff and STRPM, the proposed framework also performs better. The observation validates that modeling the precipitation system with a global trend and local stochastics is better than modeling the whole system as stochastics.

To investigate how the prediction performance changes against the lead time, we depict the curves in Figure~\ref{fig:all_res}. We observe that as the lead time increases, the performance of all method decreases, because the uncertainty for prediction is enlarged. However, the performance of methods with our framework is always better than that without it. This again validates the effectiveness of the proposed framework.

In Figure~\ref{fig:sample1}, we show and compare the results of all the methods for a precipitation event. We observe that the prediction of SimVP is blurry, as it does not model the local stochastics. MCVD, Prediff and STRPM deliver better visual details than SimVP, but the positions of green and yellow parts are less accurate. This is because these methods model the intact system in a stochastic generation way, where the freedom of generation is too high to maintain the accuracy. When equipped SimVP  into the proposed DiffCast framework, not only realistic  appearance details are produced, but also the positions are very accurate. At the 100 minute prediction, we can see that DiffCast\_SimVP accurately predicts two closing rain belts of yellow, but none of the comparison methods achieves this. The observations again validate the superiority of our framework.


\subsection{Analysis and Discussions}


\textbf{Why is the deterministic loss necessary?}
Our framework has two loss functions, namely the deterministic loss and denoising loss.
We can see from Equation (\ref{eq:l_ep}) that the denoising loss is related to all the parameters $\theta_1, \theta_2$ and $\theta_3$. Hence, by optimizing the denoising loss, we can also update the backbone predictor. Then, one would like to ask whether the deterministic loss is really necessary.

\begin{figure}[t]
    \centering
    \includegraphics[width=1\linewidth, trim=30 40 30 30, clip]{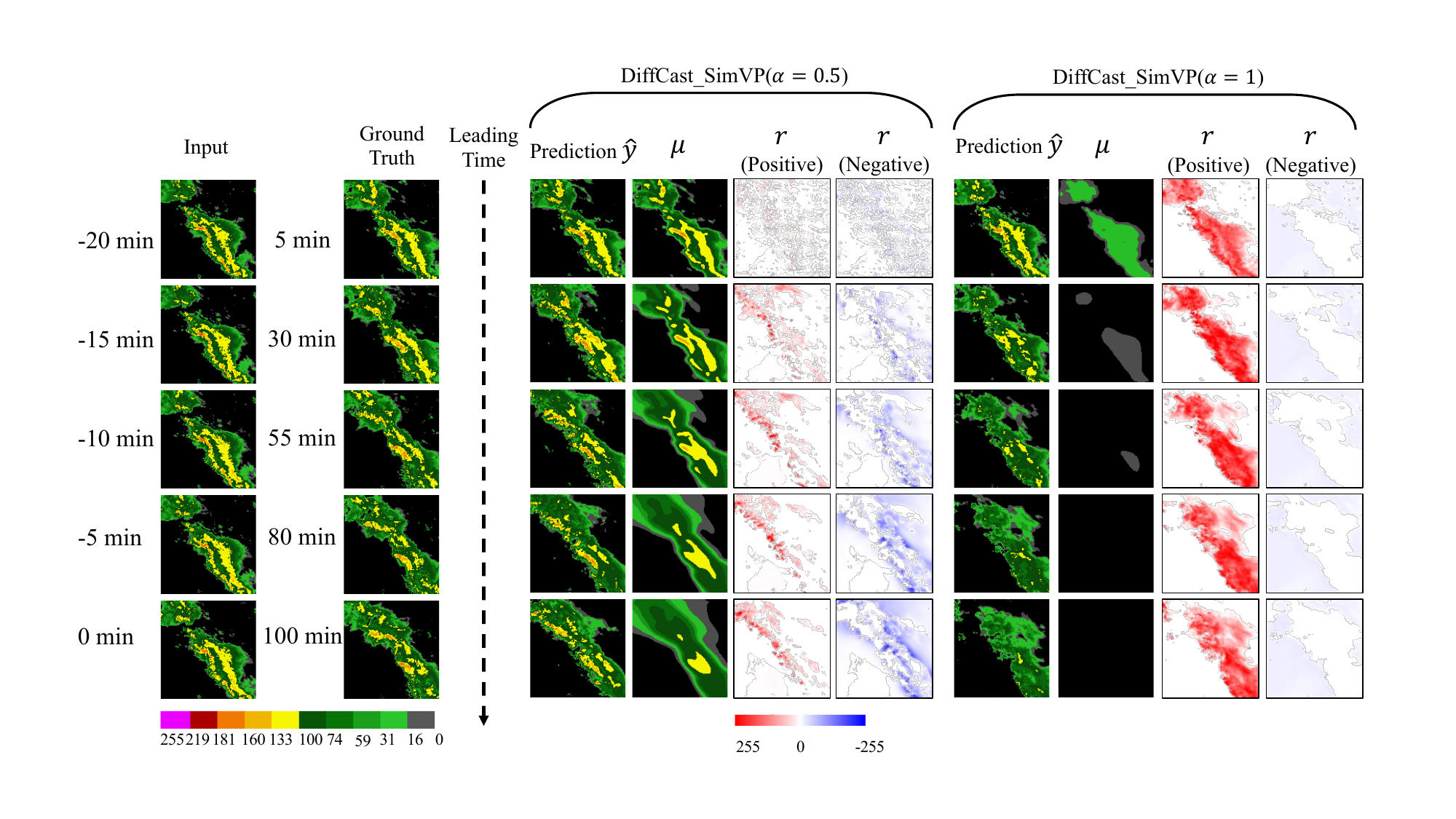}
    \vspace{-4ex}
    \caption{An illustration example of the prediction $\hat{y}$, $\mu$ and residual $r$ with or without deterministic loss for DiffCast\_SimVP.}
    \label{fig:dloss_demo}
    \vspace{-2ex}
\end{figure}

\begin{figure}[t]
  \centering
  \begin{subfigure}{0.45\linewidth}
    \includegraphics[width=1.1\columnwidth,trim=0 10 30 20, clip]{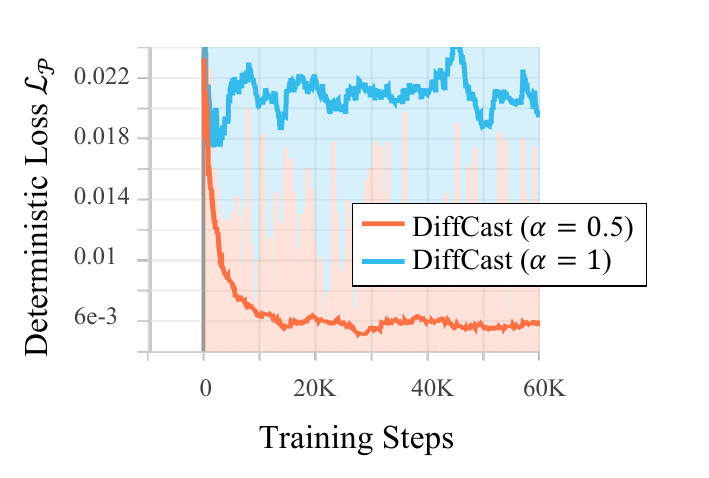}
    \caption{Deterministic Loss $\mathcal{L}_{\mathcal{P}}$}
    \label{fig:short-a}
  \end{subfigure}
  \hfill
  \begin{subfigure}{0.45\linewidth}
    \includegraphics[width=1.1\columnwidth,trim=10 5 10 2, clip]{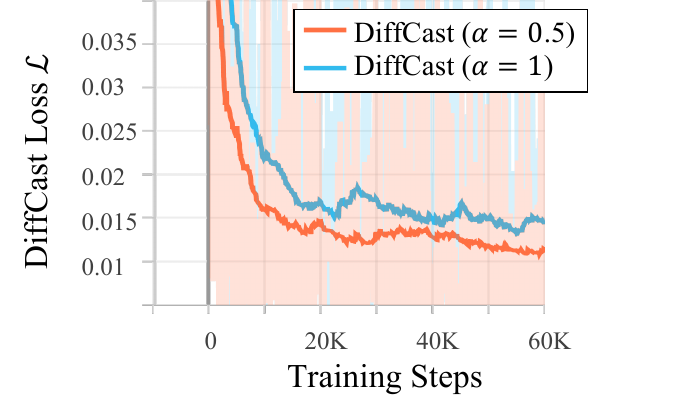}
    \caption{Overall Loss $\mathcal{L}$}
    \label{fig:short-b}
  \end{subfigure}
  \vspace{-2ex}
  \caption{Analysis of loss trends.}
  \label{fig:loss}
  \vspace{-4ex}
\end{figure}

To investigate the question, we evaluate DiffCast\_SimVP with different weight factor $\alpha$ in Equation (\ref{eq:loss_all}) and show the results in Table~\ref{tab:d_loss}. Note that DiffCast\_SimVP degenerates into SimVP when $\alpha=0$. We observe that SimVP even performs better than DiffCast\_SimVP without deterministic loss, i.e., $\alpha=1$.
To further understand the reason, we use an example to depict the prediction, $\mu$, positive and negative residuals $r$ with and without deterministic loss ${\cal L}_p$ in Figure~\ref{fig:dloss_demo}. 
We find that when both losses are used (\ie, $\alpha=0.5$), the prediction is accurate and with realistic details. In this case the $\mu$ indeed generates a global trend without details, and residual parts account for making up the details. However, if we remove the deterministic loss (\ie, $\alpha =1$), the prediction is with realistic details but not accurate. 
Moreover, in this case the $\mu$ plays a very minor role while residual becomes the main component. This is easy to understand. When the deterministic loss is removed (\ie, $\alpha=1$), our DiffCast cannot nicely decompose the precipitation system into a global trend and local stochastics, which can be validated by Figure~\ref{fig:short-a} where the deterministic loss never decreases. 
Instead, in this case the DiffCase degenerates into a model that regards the whole system as stochastics. Hence, its prediction is with realistic details but not accurate. Moreover, we can see from Figure~\ref{fig:short-b} that the deterministic loss enhances both the convergence speed and convergence quality for our DiffCast framework.


\textbf{Why do we apply end-to-end training rather than a two-stage training?} 
Our DiffCast framework is trained in an end-to-end manner with both the deterministic and stochastic components. One interesting question is whether it is really better than a two-stage strategy. In two-stage strategy, we first train a deterministic backbone, then equip it into our framework and keep it frozen. In this case, our framework degenerates into a Predictor-Corrector paradigm, where the diffusion model works as a corrector to the prediction of backbone.

\begin{table}[t]
\caption{Analysis of our end-to-end training on SEVIR}
\vspace{-2ex}
    \centering
    \resizebox{0.95\columnwidth}{!}{
    \begin{tabular}{ccccccc}
    \toprule
         Method&  CSI&  CSI-pool4&  CSI-pool16&  HSS&    LPIPS& SSIM\\
    \midrule
         DiffCast\_SimVP&  \textbf{0.3077}& \textbf{ 0.4122}& \textbf{ 0.5683}&  \textbf{0.4033}&   \textbf{ 0.1812}& \textbf{0.6354}\\
         DiffCast\_simvp-frozen&  0.2739&  0.3807&  0.5417&  0.3563&   0.1948& 0.6315\\
         \hline
         DiffCast\_Earthformer&  \textbf{0.2823} &  \textbf{0.3868} &  \textbf{0.5362} &   \textbf{ 0.3623} & \textbf{ 0.1818 }&\textbf{0.6420} \\
         DiffCast\_Earthformer-frozen& 0.2622 &0.3676  &0.5183  & 0.3400   &0.1776  &0.6341\\
\hline
         DiffCast\_PhyDnet&  \textbf{0.2757}&  \textbf{0.3797}&  \textbf{0.5296}& \textbf{ 0.3584}&   \textbf{0.1845}& \textbf{0.6320}\\
 DiffCast\_PhyDnet-frozen)& 0.2603& 0.3715& 0.5469& 0.3360&  0.1939&0.6230\\
 \bottomrule
    \end{tabular}
    }
    
    \label{tab:e2e_training}
    \vspace{-1ex}
\end{table}


Table~\ref{tab:e2e_training} shows the results. We find that the end-to-end way is absolutely better than the two-stage manner. This is because when trained in end-to-end manner, the deterministic and stochastic component can interplay with other. For example, the denoising loss of stochastic part will also produce gradients for the update of  deterministic backbone. As a result, the backbone may become better. 
\begin{table}[t]
    \caption{The ablation of DiffCast with respect to GlobalNet $\mathcal{G}_{\theta_3}$ on the SEVIR dataset.}
    \vspace{-2ex}
    \centering
     \resizebox{0.95\columnwidth}{!}{
    \begin{tabular}{ccccccc}
    \toprule
         Method&  CSI&  CSI-pool4&  CSI-pool16&  HSS&   LPIPS&SSIM\\
    \midrule
             DiffCast\_SimVP
&  \textbf{0.3077}&  \textbf{ 0.4122}&  \textbf{ 0.5683}&    \textbf{0.4033}&  \textbf{ 0.1812}&\textbf{0.6354}\\
         DiffCast\_SimVP w/o $\mathcal{G}$ &0.2719  &0.3729  &0.5471  &0.3522    &0.2135  &0.6315\\

    \hline
         DiffCast\_Earthformer& \textbf{0.2823} &  \textbf{0.3868} &  \textbf{0.5362} &    \textbf{ 0.3623} &  \textbf{ 0.1818 }&\textbf{0.6420} \\
         DiffCast\_Earthformer
w/o $\mathcal{G}$&0.2558  &0.3442 &0.4811 &0.3298 &0.1918 &0.6018\\
\hline
 DiffCast\_PhyDnet& \textbf{0.2757}& \textbf{0.3797}& \textbf{0.5296}&  \textbf{ 0.3584}& \textbf{0.1845}&0.6320\\
 DiffCast\_PhyDnet w/o $\mathcal{G}$ & 0.2648  &0.3641 &0.5049 &0.3411 &0.1760 &\textbf{0.6568}\\
    \bottomrule
    \end{tabular}
    }

    \label{tab:gnet}
    \vspace{-3ex}
\end{table}

\textbf{Is our GlobalNet $\mathcal{G}_{\theta}$ effective?}
We show the ablation study on the designed GlobalNet in Table~\ref{tab:gnet}. We can see that removing GlobalNet degrades the performance, which indicates its effectiveness.



\section{Conclusion}
In this paper, we propose an unified and flexible framework for precipitation nowcasting based on residual diffusion model, which nicely decomposes and models the system evolution with a global deterministic trend component and a local stochastic component. Extensive experiments on four real-world datasets verifies the effectiveness of the proposed framework.

\section*{Acknowledgement}
This work was supported in part by NSFC under Grants 62376072, 62272130, and in part by Shenzhen Science and Technology Program No. KCXFZ20211020163403005 and in part by Science and technology innovation team project of Guangdong Meteorological Bureau (GRMCTD202104), Innovation and Development Project of China Meteorological Administration (CXFZ2022J002), Shenzhen Hong Kong Macao science and technology plan project (SGDX20210823103537035).

%% file: sec/X_suppl.tex
\clearpage
\setcounter{page}{1}
\maketitlesupplementary

\section{Datasets Detials}
\label{sec:suppl_data}
\textbf{SEVIR}, the Storm EVent ImagRy (SEVIR)~\cite{veillette2020sevir} is an annotated, curated and spatio-temporally aligned dataset across five multiple data types including visible satellite imagery, infrared satellite imagery (mid-level water vapor and clean longwave window), NEXRAD radar mosaic of VIL(vertically integrated liquid mosaics) and ground lightning events. In this paper, we focus on the short term weather forecasting task and select all the radar mosaics of VIL as the main data. The dataset contains 20393 weather events from multiple sensors in 2017-2020. Each event consists of a 4-hour length sequence of images sampled in 5 minute steps covering 384 km$\times$384 km patches sampled at locations throughout the continental U.S.. As our task is to predict the future VIL up to 20 frames (100 min) given 5 observed frames (25 min), we follow~\cite{gao2022earthformer} to sample the 25 continuous frames with stride = 12 in every event and split the dataset into training, validation and test sets with the time point January 1, 2019 and June 1, 2019, respectively. The frames are rescaled back to the range 0-255 and binarized at thresholds [16,74,133,160,181,219] to calculate the CSI and HSS following original settings in~\cite{veillette2020sevir}.

\textbf{MeteoNet}~\cite{2020meteo} is a multimodel dataset including full time series of satellite and radar images, weather models and ground observations.
It covers geographic areas of 550 km$\times$550 km in the northwestern quarter of France and a span over three
years, and records every 6 min from 2016 to 2018. 
Like the SEVIR, we split the radar sequence from 2016 to 2018 into training, validation and test sets with the time point January 1, 2018 and June 1, 2018, respectively. Then, we apply Algorithm~\ref{alg:data_filter} to filter precipitation events with a stride-20 sliding window to reduce the noise in the data. Note that a mean pixel threshold $T\_{pixel}$ is used as a filter to precipitation events.
The data range of frames in MeteoNet is set to [0-70] and the thresholds are set to [12, 18, 24, 32] following~\cite{2020meteo} for the CSI and HSS evaluation.

\textbf{Shanghai\_Radar}~\cite{chen2020deep} is a dataset contains continuous radar echo frames generated by volume scans in intervals of approximately 6 minute from October 2015 to July 2018 in Pudong, Shanghai. Every radar echo map covers 501 km$\times$501 km area.  We follow~\cite{chen2020deep} to preprocess the echo sequence and also apply Algorithm~\ref{alg:data_filter} to filter 25-frame weather event datasets. The data range of frames in Shanghai Radar is set to [0-70] and the thresholds are set to [20, 30, 35, 40] following~\cite{2020meteo} for the CSI and HSS computation.

\textbf{CIKM} is a radar dataset from CIKM AnalytiCup 2017 Competition, recording precipitation samples in 101 km$\times$101km area of Guangdong, China. Each sample settles 15 historical radar echo maps as a sample in which the time interval between two consecutive maps is 6-minute. We follow~\cite{luo2022reconstitution} to process the dataset to pad each echo map into $128 \times 128$ and follow the original setting to split training, validation and test sets.
We transform the pixel in each frame to  the reflectivity of [0,76] dBZ and use the thresholds [20,30,35,40] to compute the CSI and HSS.

The lengths of event sequences in each dataset are set to 25 frames except for the CIKM dataset with 15 frames. 
Compared to most of the existing studies, which aim to make an hour prediction (\eg, 10 frames with a 6-minute interval), our tasks (except for CIKM dataset) are for the forecast in two hours (\ie 20 frames) in this paper, which are more challenging.
Although some recent studies attempt to achieve two hours prediction by frame interpolation (\eg, predicting 10 frames with a 12-minute interval), this trick simplifies the complexity of precipitation dynamics and results in a degrading temporal resolution for prediction.

\begin{algorithm}
\caption{Weather Event Filtering}
\label{alg:data_filter}
\begin{algorithmic}[1]
\State Given continuous frames $s$, pixel threshold $T_{pixel}$
\State $i \gets 10$
\State  $L_{in},L_{out} \gets 5,20$
\State $event\_set \gets \{\}$ 
\While{$i + L_{out} < \text{Len}(s)$}
    \If{$\text{Mean}(s[i]) > T_{pixel}$}
        \State $event \gets s[i-L_{in} : i+L_{out}]$
        \State $event\_pixel \gets \sum_{frame \in event} \text{Mean}(frame)$
        \If{$event\_pixel >= (L_{in} + L_{out})T_{pixel}/2$}
            \State Add event to $event\_set$.
            \State $i \gets i + L_{out}$
            \State Continue
        \EndIf
    \EndIf
    \State $i \gets i + 1$
\EndWhile
\State Return $event\_set$
\end{algorithmic}
\end{algorithm}

\section{DiffCast: Implementation Details}
In this section, we will give a detailed description of the implementation for \Change{ DiffCast's main architecture and its training and inference process, as well as our experimental settings.}

\Add{\textbf{Architecture of DiffCast}.
We have described the main architecture of the DiffCast model in section \ref{sec:unet_block}. 
Here, we present our detailed implementation of the Temp-Attn Block and GlobalNet, which are mainly composed of temporal attention operator\cite{chang2021mau, tan2023temporal} and ConvGRU operator\cite{shi2017deep,yang2022diffusion}, respectively, as summarized in Table \ref{tab:network}.
}

\begin{table}
    \centering
    \caption{Detailed implementation of our Temp-Attn Block and GlobalNet.}
    \vspace{-2ex}
        \resizebox{1\columnwidth}{!}{
    \begin{tabular}{c|cc}
    \toprule
         \multicolumn{3}{c}{Temp-Attn Block}\\
    \hline
         ResBlock$\times$2&  2$\times$[Conv3x3 + GroupNorm8+ SiLU] + Conv3x3& Res Operator\\
         Temporal Attention&  Conv5x5 (Spatial) + Conv1x1 (Temporal)+FC& Attention Operator\\
         Down/Upsampler&  Conv1x1& \\
    \bottomrule
    \toprule
         \multicolumn{3}{c}{GlobalNet}\\
    \hline
         ResBlock$\times$4&  2$\times$[Conv3x3 + GroupNorm8+ SiLU] + Conv3x3& Res Operator\\
         ConvGRU$\times$4&  Conv3x3 + Conv3x3;  HiddenState& GRU Operator\\
         Downsampler$\times$4&  Conv1x1& \\
    \bottomrule
    \end{tabular}
    }
    \label{tab:network}
    \vspace{-2ex}
\end{table}

\begin{algorithm}

    \caption{Training of The Framework}
    \begin{algorithmic}[1]
    \While{not converged}
        \State Sampling a sequence $(x,y) \sim \mathcal{D},$ where $len(x) = L_{in}, len(y) = L_{out}$ 
        \State Making basic prediction $\mu = \mathcal{P}_{\theta_{1}}(x),$ where $len(\mu) = L_{out}$
        \State Building residual sequence $r$ following Eq.~(\ref{eq:residual})
        \State Grouping segments $s_j$ from $r$ following Eq.~(\ref{eq:s_j-1})
        \State Extracting global hidden state $h$ following Eq.~(\ref{eq:hs})
        \State Sampling diffusion step $t\sim \mathcal{U}(0,...,T)$
        \State $\mathcal{L}_{\epsilon} \gets 0$
        \While{$j < \lceil\frac{L_\text{out}}{K}\rceil$}
            \State $\epsilon \sim \mathcal{N}(0, I)$
            \State Disturbing $s_j$ to $s_j^t$ following Eq.~(\ref{eq:q0})
            \State Getting denoising loss $\mathcal{L}_{\epsilon}^j$ following Eq.~(\ref{eq:l_ep})
            \State $\mathcal{L}_{\epsilon} = \mathcal{L}_{\epsilon} + \mathcal{L}_{\epsilon}^j$
        \EndWhile
        \State Computing deterministic loss $\mathcal{L}_{\mathcal{P}}$ following Eq.~(\ref{eq:loss_mu})
        \State Computing final loss $\mathcal{L}$ following Eq.~(\ref{eq:loss_all}) given $\alpha$
        \State $(\theta_1,\theta_2, \theta_3) \gets (\theta_1,\theta_2, \theta_3) - \triangledown_{(\theta_1,\theta_2, \theta_3)} \mathcal{L}$
    \EndWhile
    \end{algorithmic}
    \label{alg:train}
\end{algorithm}

\begin{algorithm}
    \caption{Inference of The Framework}
    \begin{algorithmic}[1]
        \State Given initial frames $x$
        \State Making basic prediction $\mu = \mathcal{P}_{\theta_{1}}(x)$
        \State Extracting global hidden state $h$ following Eq.~(\ref{eq:hs})
        \State $j \gets 0, \hat{s}_{j-1} \gets 0$
        \While{$j < \lceil\frac{L_\text{out}}{K}\rceil$}
            \State $s_{j}^T \sim \mathcal{N}(0,1)$
            \While{Reverse diffusion from $t=T$ to $t=1$}
                \State $\epsilon \sim \mathcal{N}(0, I)$
                \State Estimating target noise $\epsilon_{\theta_2}$ following Eq.~(\ref{eq:denoising})
                \State Recovering $s_j^{t-1}$ from $s_j^t$ following Eq.~(\ref{eq:yt-1})
            \EndWhile
            \State Getting current residual segment $\hat{s}_j$
        \EndWhile
        \State Computing target frames $\hat{y}$ following Eq.~(\ref{eq:hat_y})
    \end{algorithmic}
\label{alg:inference}   
\end{algorithm}

\textbf{Training and Inference}. The DiffCast is trained with an end-to-end manner as shown in Figure~\ref{fig:framework} (b), where the base deterministic predictor and residual diffusion model are optimized within the same training iteration. The complete training procedure is summarized in Algorithm~\ref{alg:train}. 
In the inference phase, the framework also utilizes the base predictor to estimate the global trend and then apply the diffusion model to generate the residual segments autogressively. The final prediction is obtained by combining the two components. The inference procedure is summarized in Algorithm~\ref{alg:inference}.

\textbf{Experimental details}
All experiments are conducted on a computer with NVIDIA A6000 GPU (48G memory) and all models, including DiffCast equipped with various backbones and single backbones, can fit in a single GPU. As for the implementation of various backbones, we easily rebuild the most backbones from OpenSTL~\cite{tan2023openstl} library to adapt with DiffCast. 
We construct the GTUNet with a hierarchical UNet architecture with temporal attention blocks. This structure is composed of four up/down layers, with a hidden size of 64, and is subsequently upscaled/downscaled by factors of 1,2,4,8, respectively.
Despite the increased number of parameters (\eg, from 20.13MB to 66.40 MB for DiffCast\_MAU) and higher training costs (\eg, from 18 hours at 12 batch size to 23 hours at 6 batch size for 30K iterations) associated with the DiffCast framework, we can attain a great paramount for modeling the distribution of stochastic temporal evolution. Furthermore, the DiffCast framework can expedite the forecast by utilizing optimization techniques such as DDIM, DPM-solver \etc. This capability  can fully meet the requirements of short-term precipitation forecasting task in real-world scenarios (\eg, 17 seconds to produce 20-frame forecasts), enabling real-time predictions that are both more accurate and realistic.

\section{Additional Analysis}
In this section, we give an extra analysis on the design of framework loss, the computational complexity and the hyperparameter $K$.

\Add{\textbf{w/o stochastic Loss}}. We decompose the determinism and local stochastics in precipitation evolution and model them with a deterministic component and a residual diffusion component, respectively. Different from other two-stage frameworks, we train the overall framework in an end-to-end manner to simulate the interplay between the determinism and uncertainty, which indicate that the gradient from stochastic diffusion denoising loss can also optimize the deterministic backbone. In Table~\ref{tab:mu}, we compare the performance between the pure deterministic backbones and the intermediate output $\mu$ from deterministic component in DiffCast. The results show that the stochastic loss indeed leads to a positive optimization on most of the deterministic backbones.
\begin{table}[t]
    \centering
    \caption{Analysis of performance for backbones with different optimization strategies on SEVIR.}
    \resizebox{0.95\columnwidth}{!}{
    \begin{tabular}{ccccccc}
    \toprule
         Deterministic Method&  CSI&  CSI-pool4&  CSI-pool16&  HSS&    LPIPS& SSIM\\
         \midrule
         SimVP&  0.2662 &  0.2844 &  0.3452 &  0.3369 &    0.3914 & 0.6570\\
         DiffCast\_Simvp-$\mu$&  \textbf{0.2690}&  \textbf{0.2828}&  \textbf{0.3134}&  \textbf{0.3320}&    \textbf{0.3961}& \textbf{0.6728}\\
         \hline
           Earthformer&\textbf{0.2513}&\textbf{0.2617}&\textbf{0.2910}&\textbf{0.3073}&\textbf{0.4140}&\textbf{0.6773}\\
         DiffCast\_Earthformer-$\mu$&0.2490 &0.2579 &0.2834 &0.3040  &0.4391 &0.6685\\
         \hline
 MAU& 0.2463& 0.2566& 0.2861& 0.3004 &\textbf{0.3933}&0.6361\\
 DiffCast\_MAU-$\mu$ &  \textbf{0.2542}& \textbf{0.2848}& \textbf{0.3157}& \textbf{0.3361}&0.3929&\textbf{0.7028}\\
 \hline
 ConvGRU& 0.2560& 0.2685 & 0.3005 & 0.3124  & 0.3785 &\textbf{0.6764} \\
 DiffCast\_ConvGRU-$\mu$&  \textbf{0.2635}&\textbf{0.2873}&\textbf{0.3197}& \textbf{0.3350}& \textbf{0.3860}&\textbf{0.6818}\\
         \hline
         PhyDNet&0.2560	&0.2685 	&0.3005 	&0.3124 		&0.3785 	&0.6764 \\
          DiffCast\_PhyDNet-$\mu$&\textbf{0.2659} &\textbf{0.2785} &\textbf{0.3105} &\textbf{0.3252}  &\textbf{0.3748} &\textbf{0.6811}\\
        \bottomrule
        \end{tabular}
    }

    \label{tab:mu}

\end{table}

\begin{figure}[t]
    \centering
    \includegraphics[width=1\linewidth, trim=0 30 0 0, clip ]{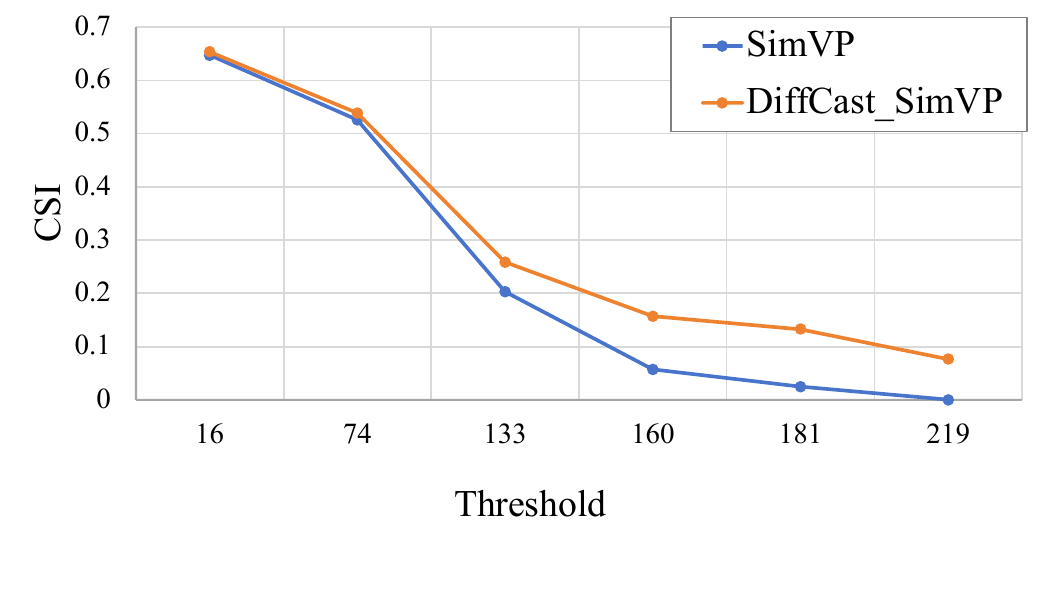}
    \caption{CSI with threshold}
    \label{fig:csi_with_threshold}
\end{figure}

\begin{figure}[t]
    \centering
    \includegraphics[width=1\linewidth]{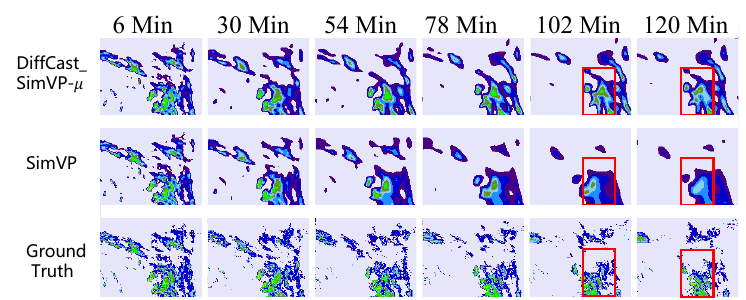}
    \caption{\Add{Qualitative results of SimVP w/o stochastic loss.}}
    \label{fig:mu_simvp}
\end{figure}

We point out that the conventional deterministic methods always under-estimate the high-value echoes with the increasing lead time (shown in Figure~\ref{fig:intro}). 
To further investigate this, we show in Figure~\ref{fig:csi_with_threshold} the performance of DiffCast\_SimVP, in terms of different thresholds. We observe that with the stochastic modeling the high-value echoes indeed can be more accurately maintained and predicted. 
\Add{Additionally, we show the qualitative results of SimVP w/o stochastic diffusion loss in Figure \ref{fig:mu_simvp}. The results indicate that DiffCast$\_$SimVP-$\mu$ alleviates the echo value fading away issue compared to SimVP, which implies that stochastic objectives indeed help optimize the deterministic model.}


\textbf{Complexity and Hyperparameter $K$}.
\Add{In Table \ref{tab:size}, we report the tradeoff between model size, memory and time cost conditioned on different segment length based on our experimental setting with batchsize=4 for 30K training iterations. $K$ is selected from \{$2,4,5,10$\} on validation set and $K$=5 delivers the best tradeoff. 
There are more requirements for memory compared with base predictor but it is acceptable in our practical application.
It is notable that the model size is not influenced by $K$.
}

\begin{table}
    \centering
    \caption{\Add{Complexity analysis and hyperparameter $K$.}}
    \vspace{-2ex}
    \resizebox{1\columnwidth}{!}{
    \begin{tabular}{c|clcccc}
    \toprule
         &   &&\multicolumn{2}{c}{Training}&  \multicolumn{2}{c}{Inference}\\
         &  Model Size &CSI&  Memory&  Time Cost&  Memory& Time Cost\\
         \hline
         MAU&  20.13M &0.2463&  21759MB&  14.5h&  2297MB& 0.29s\\
         DiffCast\_MAU(K=2)&  66.38M &0.2638&  43815MB&  22.8 h&  3881MB& 42s\\
         DiffCast\_MAU(K=4)&  66.39M &0.2697&  35471MB&  20.5 h&  3731MB& 21s\\
         DiffCast\_MAU(K=5)&  66.40M &\textbf{0.2716}&  33791MB&  19.5 h&  3815MB& 16s\\
         DiffCast\_MAU(K=10)&  66.43M &0.2548&  30443MB&  18.5h&  4065MB& 8s\\
    \bottomrule
    \end{tabular}
    }
    \label{tab:size}
      \vspace{-2ex}
\end{table}

\section{More Qualitative Results}

\begin{figure*}[t]
    \centering
    \resizebox{0.85\linewidth}{!}{
    \includegraphics[width=1\linewidth,trim=0 0 0 0, clip]{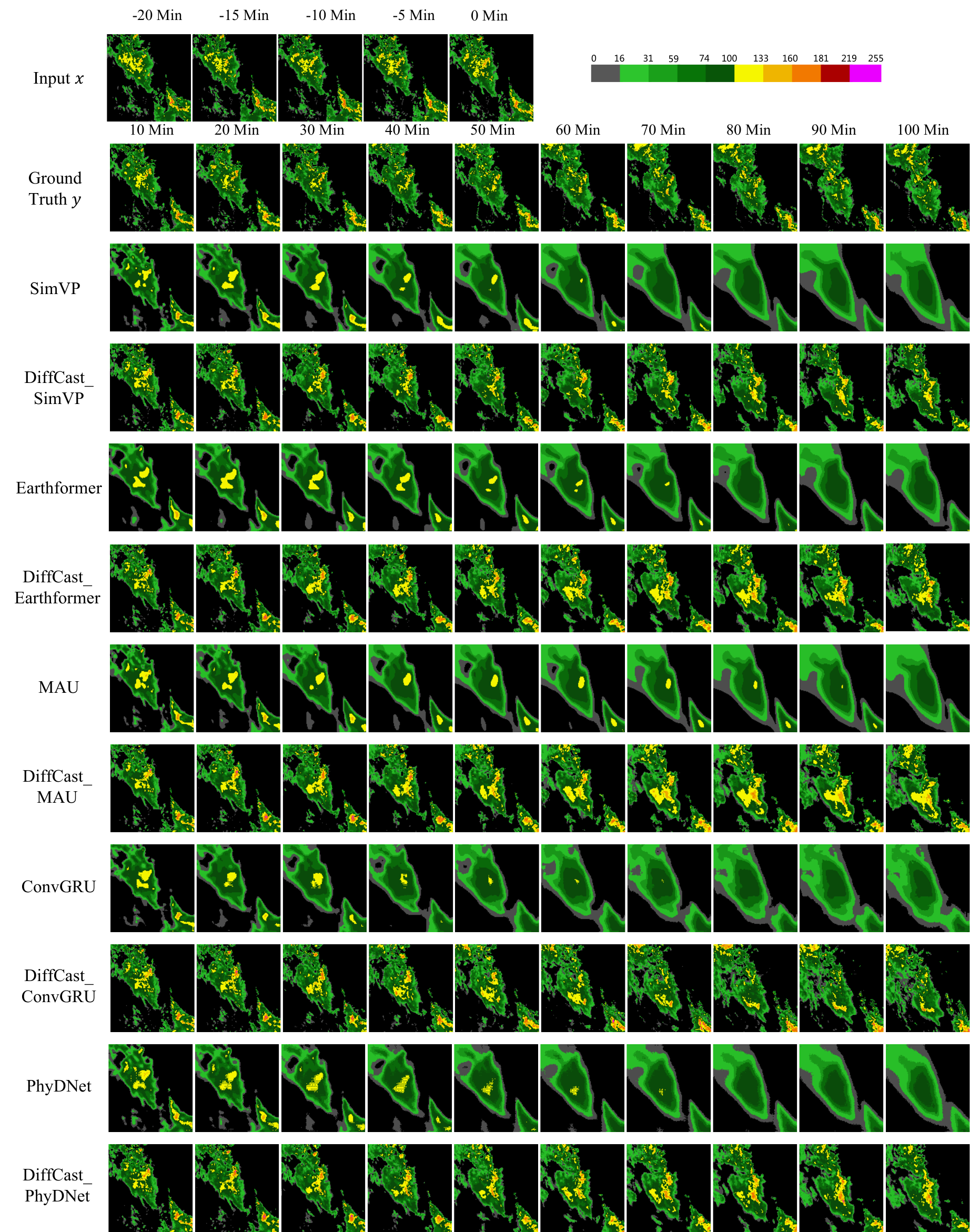}
    }
    \caption{Prediction examples on the SEVIR.}
    \label{fig:sevir}
\end{figure*}

\begin{figure*}[t]
    \centering
    \resizebox{0.85\linewidth}{!}{
    \includegraphics[width=1\linewidth,trim=0 0 0 0, clip]{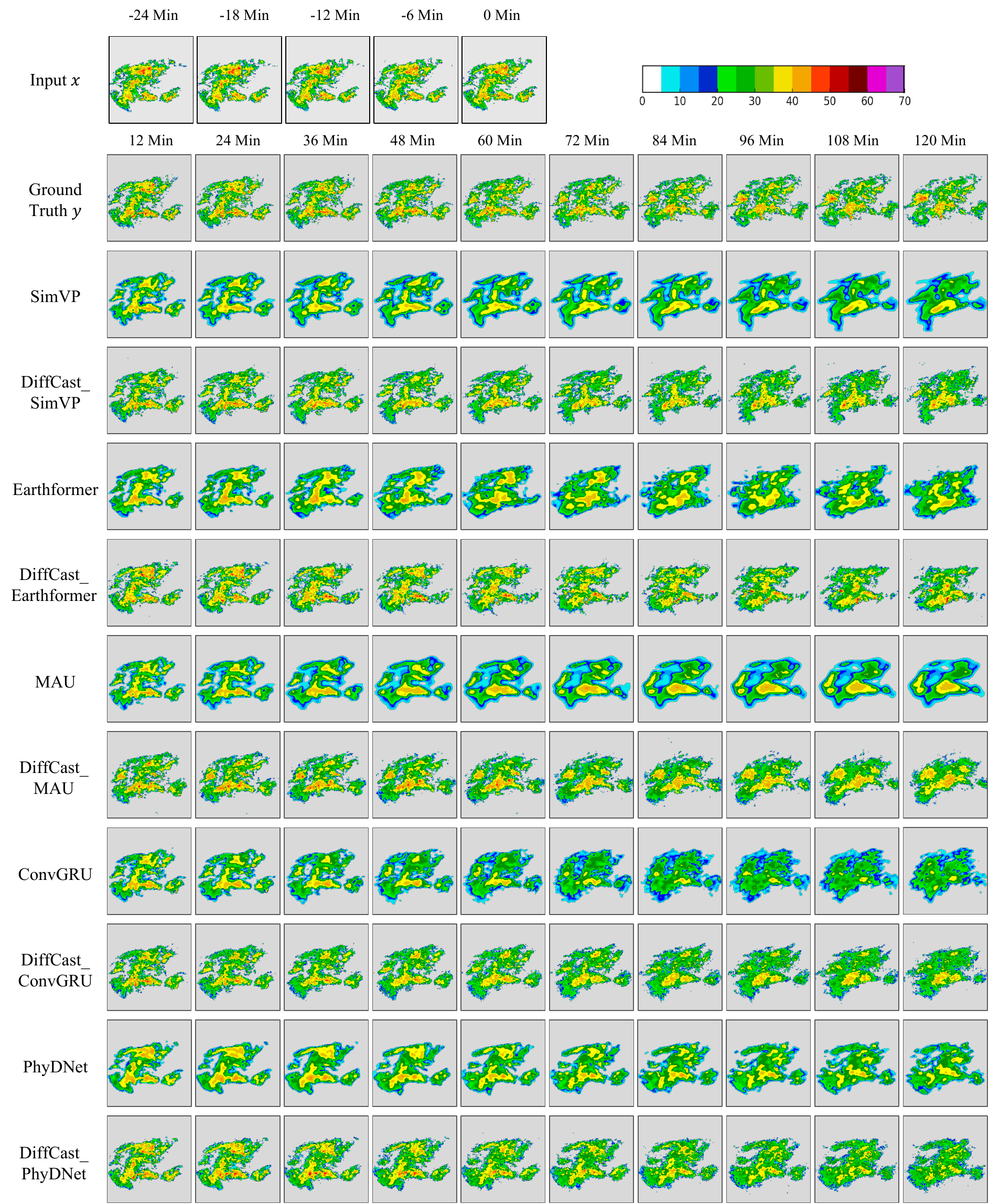}
    }
    \caption{Prediction examples on the Shanghai Radar.}
    \label{fig:shanghai}
\end{figure*}

\begin{figure*}[t]
    \centering
    \includegraphics[width=1\linewidth]{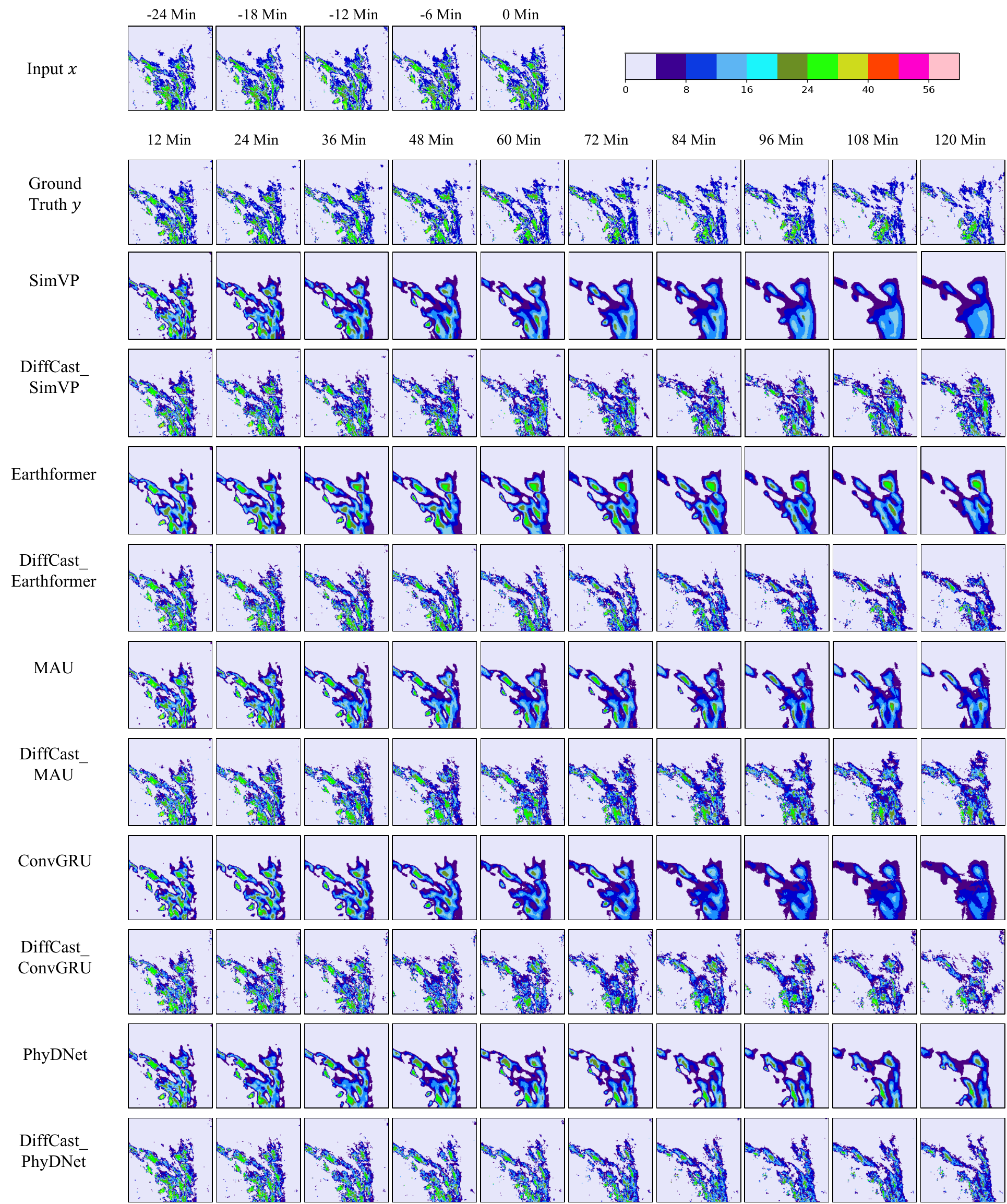}
    \caption{Prediction examples on the MeteoNet.}
    \label{fig:meteo}
\end{figure*}

\begin{figure*}[t]
    \centering
    \includegraphics[width=1\linewidth]{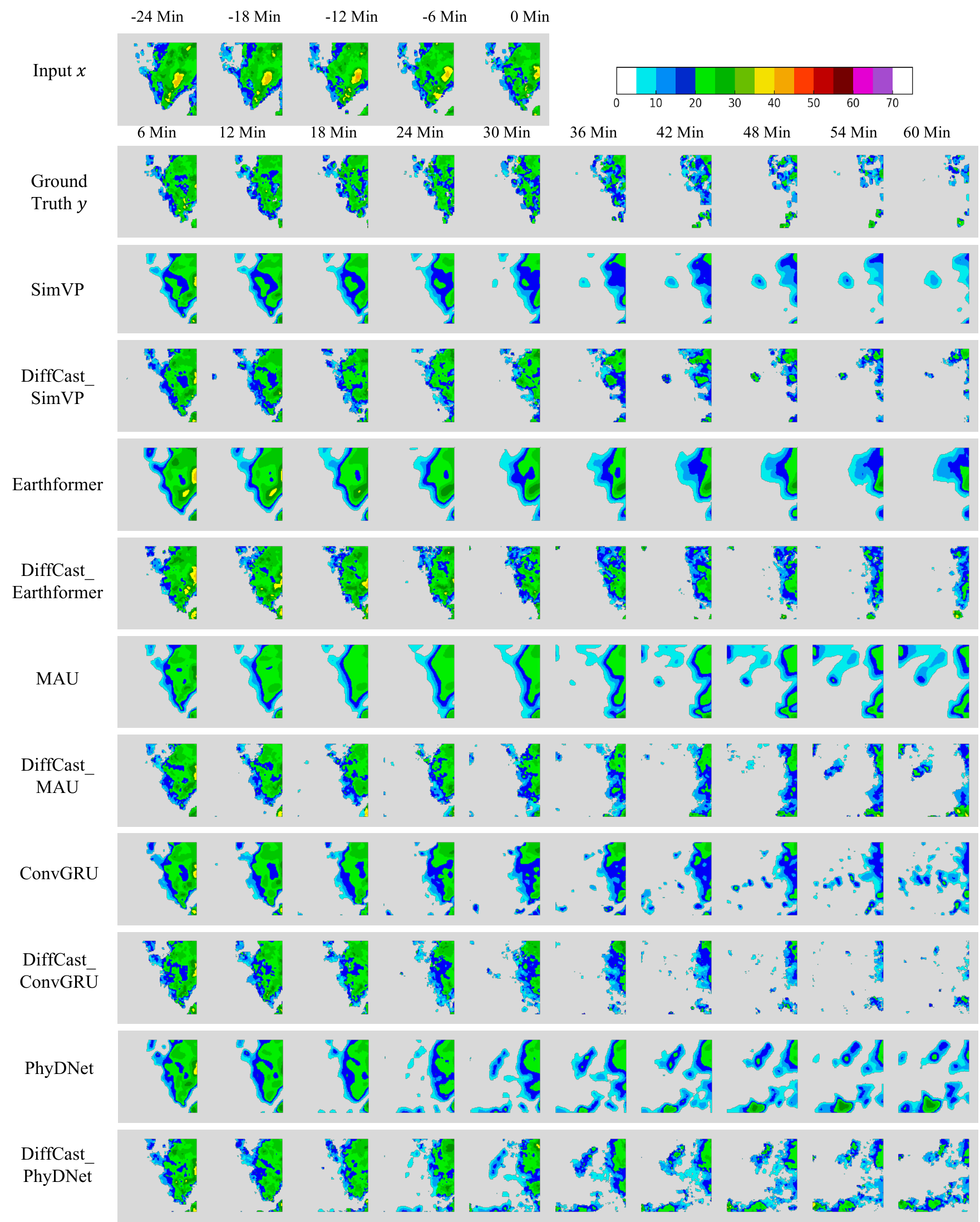}
    \caption{Prediction examples on the CIKM dataset.}
    \label{fig:cikm}
\end{figure*}

In this section, we show more illustrative examples on different datasets to compare our DiffCast with baseline methods.
As shown in Figure~\ref{fig:sevir},~\ref{fig:meteo},~\ref{fig:shanghai},~\ref{fig:cikm}, all deterministic backbones deliver blurry results after 60 minutes, with a phenomenon of high-value echoes and details fading away.
However, when equipped into our DiffCast framework, all the prediction results of the backbones are consistently enhanced, where the forecast images are not blurry anymore, and the high-value echoes and details are carefully preserved.  All the observations validate the effectiveness of the proposed DiffCast framework.


%% file: main.bbl
\begin{thebibliography}{39}
\providecommand{\natexlab}[1]{#1}
\providecommand{\url}[1]{\texttt{#1}}
\expandafter\ifx\csname urlstyle\endcsname\relax
  \providecommand{\doi}[1]{doi: #1}\else
  \providecommand{\doi}{doi: \begingroup \urlstyle{rm}\Url}\fi

\bibitem[Babaeizadeh et~al.(2018)Babaeizadeh, Finn, Erhan, Campbell, and Levine]{babaeizadeh2018stochastic}
Mohammad Babaeizadeh, Chelsea Finn, Dumitru Erhan, Roy Campbell, and Sergey Levine.
\newblock Stochastic variational video prediction.
\newblock In \emph{International Conference on Learning Representations}, 2018.

\bibitem[Bai et~al.(2022)Bai, Sun, Zhang, Song, and Chen]{bai2022rainformer}
Cong Bai, Feng Sun, Jinglin Zhang, Yi Song, and Shengyong Chen.
\newblock Rainformer: Features extraction balanced network for radar-based precipitation nowcasting.
\newblock \emph{IEEE Geoscience and Remote Sensing Letters}, 19:\penalty0 1--5, 2022.

\bibitem[Chang et~al.(2021)Chang, Zhang, Wang, Ma, Ye, Xinguang, and Gao]{chang2021mau}
Zheng Chang, Xinfeng Zhang, Shanshe Wang, Siwei Ma, Yan Ye, Xiang Xinguang, and Wen Gao.
\newblock Mau: A motion-aware unit for video prediction and beyond.
\newblock \emph{Advances in Neural Information Processing Systems}, 34:\penalty0 26950--26962, 2021.

\bibitem[Chang et~al.(2022)Chang, Zhang, Wang, Ma, and Gao]{chang2022strpm}
Zheng Chang, Xinfeng Zhang, Shanshe Wang, Siwei Ma, and Wen Gao.
\newblock Strpm: A spatiotemporal residual predictive model for high-resolution video prediction.
\newblock In \emph{Proceedings of the IEEE/CVF Conference on Computer Vision and Pattern Recognition}, pages 13946--13955, 2022.

\bibitem[Chen et~al.(2020)Chen, Cao, Ma, and Zhang]{chen2020deep}
Lei Chen, Yuan Cao, Leiming Ma, and Junping Zhang.
\newblock A deep learning-based methodology for precipitation nowcasting with radar.
\newblock \emph{Earth and Space Science}, 7\penalty0 (2):\penalty0 e2019EA000812, 2020.

\bibitem[Dhariwal and Nichol(2021)]{dhariwal2021diffusion}
Prafulla Dhariwal and Alexander Nichol.
\newblock Diffusion models beat gans on image synthesis.
\newblock \emph{Advances in neural information processing systems}, 34:\penalty0 8780--8794, 2021.

\bibitem[Franceschi et~al.(2020)Franceschi, Delasalles, Chen, Lamprier, and Gallinari]{franceschi2020stochastic}
Jean-Yves Franceschi, Edouard Delasalles, Micka{\"e}l Chen, Sylvain Lamprier, and Patrick Gallinari.
\newblock Stochastic latent residual video prediction.
\newblock In \emph{International Conference on Machine Learning}, pages 3233--3246. PMLR, 2020.

\bibitem[Gao et~al.(2022{\natexlab{a}})Gao, Shi, Wang, Zhu, Wang, Li, and Yeung]{gao2022earthformer}
Zhihan Gao, Xingjian Shi, Hao Wang, Yi Zhu, Yuyang~Bernie Wang, Mu Li, and Dit-Yan Yeung.
\newblock Earthformer: Exploring space-time transformers for earth system forecasting.
\newblock \emph{Advances in Neural Information Processing Systems}, 35:\penalty0 25390--25403, 2022{\natexlab{a}}.

\bibitem[Gao et~al.(2022{\natexlab{b}})Gao, Tan, Wu, and Li]{gao2022simvp}
Zhangyang Gao, Cheng Tan, Lirong Wu, and Stan~Z Li.
\newblock Simvp: Simpler yet better video prediction.
\newblock In \emph{Proceedings of the IEEE/CVF Conference on Computer Vision and Pattern Recognition}, pages 3170--3180, 2022{\natexlab{b}}.

\bibitem[Gao et~al.(2023)Gao, Shi, Han, Wang, Jin, Maddix, Zhu, Li, and Wang]{gao2023prediff}
Zhihan Gao, Xingjian Shi, Boran Han, Hao Wang, Xiaoyong Jin, Danielle Maddix, Yi Zhu, Mu Li, and Yuyang Wang.
\newblock Prediff: Precipitation nowcasting with latent diffusion models.
\newblock \emph{arXiv preprint arXiv:2307.10422}, 2023.

\bibitem[Guen and Thome(2020)]{guen2020disentangling}
Vincent~Le Guen and Nicolas Thome.
\newblock Disentangling physical dynamics from unknown factors for unsupervised video prediction.
\newblock In \emph{Proceedings of the IEEE/CVF Conference on Computer Vision and Pattern Recognition}, pages 11474--11484, 2020.

\bibitem[Harvey et~al.(2022)Harvey, Naderiparizi, Masrani, Weilbach, and Wood]{harvey2022flexible}
William Harvey, Saeid Naderiparizi, Vaden Masrani, Christian Weilbach, and Frank Wood.
\newblock Flexible diffusion modeling of long videos.
\newblock \emph{Advances in Neural Information Processing Systems}, 35:\penalty0 27953--27965, 2022.

\bibitem[Hatanaka et~al.(2023)Hatanaka, Glaser, Galgon, Torri, and Sadowski]{hatanaka2023diffusion}
Yusuke Hatanaka, Yannik Glaser, Geoff Galgon, Giuseppe Torri, and Peter Sadowski.
\newblock Diffusion models for high-resolution solar forecasts.
\newblock \emph{arXiv preprint arXiv:2302.00170}, 2023.

\bibitem[Ho et~al.(2020)Ho, Jain, and Abbeel]{ho2020denoising}
Jonathan Ho, Ajay Jain, and Pieter Abbeel.
\newblock Denoising diffusion probabilistic models.
\newblock \emph{Advances in neural information processing systems}, 33:\penalty0 6840--6851, 2020.

\bibitem[H{\"o}ppe et~al.(2022)H{\"o}ppe, Mehrjou, Bauer, Nielsen, and Dittadi]{hoppe2022diffusion}
Tobias H{\"o}ppe, Arash Mehrjou, Stefan Bauer, Didrik Nielsen, and Andrea Dittadi.
\newblock Diffusion models for video prediction and infilling.
\newblock \emph{Transactions on Machine Learning Research}, 2022.

\bibitem[Jiang et~al.(2023)Jiang, Cornman, Park, Sapp, Zhou, Anguelov, et~al.]{jiang2023motiondiffuser}
Chiyu Jiang, Andre Cornman, Cheolho Park, Benjamin Sapp, Yin Zhou, Dragomir Anguelov, et~al.
\newblock Motiondiffuser: Controllable multi-agent motion prediction using diffusion.
\newblock In \emph{Proceedings of the IEEE/CVF Conference on Computer Vision and Pattern Recognition}, pages 9644--9653, 2023.

\bibitem[Larvor et~al.(2020)Larvor, Berthomier, Chabot, Pape, Pradel, and Perez]{2020meteo}
Gwennaëlle Larvor, Léa Berthomier, Vincent Chabot, Brice~Le Pape, Bruno Pradel, and Lior Perez.
\newblock Meteonet, an open reference weather dataset, 2020.

\bibitem[Lee et~al.(2018)Lee, Zhang, Ebert, Abbeel, Finn, and Levine]{lee2018stochastic}
Alex~X Lee, Richard Zhang, Frederik Ebert, Pieter Abbeel, Chelsea Finn, and Sergey Levine.
\newblock Stochastic adversarial video prediction.
\newblock \emph{arXiv preprint arXiv:1804.01523}, 2018.

\bibitem[Leinonen et~al.(2023)Leinonen, Hamann, Nerini, Germann, and Franch]{leinonen2023latent}
Jussi Leinonen, Ulrich Hamann, Daniele Nerini, Urs Germann, and Gabriele Franch.
\newblock Latent diffusion models for generative precipitation nowcasting with accurate uncertainty quantification.
\newblock \emph{arXiv preprint arXiv:2304.12891}, 2023.

\bibitem[Luo et~al.(2022{\natexlab{a}})Luo, Li, Ye, Feng, and Ng]{luo2022experimental}
Chuyao Luo, Xutao Li, Yunming Ye, Shanshan Feng, and Michael~K Ng.
\newblock Experimental study on generative adversarial network for precipitation nowcasting.
\newblock \emph{IEEE Transactions on Geoscience and Remote Sensing}, 60:\penalty0 1--20, 2022{\natexlab{a}}.

\bibitem[Luo et~al.(2022{\natexlab{b}})Luo, Xu, Li, and Ye]{luo2022reconstitution}
Chuyao Luo, Guangning Xu, Xutao Li, and Yunming Ye.
\newblock The reconstitution predictive network for precipitation nowcasting.
\newblock \emph{Neurocomputing}, 507:\penalty0 1--15, 2022{\natexlab{b}}.

\bibitem[Mei and Patel(2023)]{mei2023vidm}
Kangfu Mei and Vishal Patel.
\newblock Vidm: Video implicit diffusion models.
\newblock In \emph{Proceedings of the AAAI Conference on Artificial Intelligence}, pages 9117--9125, 2023.

\bibitem[Ning et~al.(2023)Ning, Lan, Li, Chen, Chen, Chen, Han, and Cui]{ning2023mimo}
Shuliang Ning, Mengcheng Lan, Yanran Li, Chaofeng Chen, Qian Chen, Xunlai Chen, Xiaoguang Han, and Shuguang Cui.
\newblock Mimo is all you need: a strong multi-in-multi-out baseline for video prediction.
\newblock In \emph{Proceedings of the AAAI Conference on Artificial Intelligence}, pages 1975--1983, 2023.

\bibitem[Rasul et~al.(2021)Rasul, Seward, Schuster, and Vollgraf]{rasul2021autoregressive}
Kashif Rasul, Calvin Seward, Ingmar Schuster, and Roland Vollgraf.
\newblock Autoregressive denoising diffusion models for multivariate probabilistic time series forecasting.
\newblock In \emph{International Conference on Machine Learning}, pages 8857--8868. PMLR, 2021.

\bibitem[Ravuri et~al.(2021)Ravuri, Lenc, Willson, Kangin, Lam, Mirowski, Fitzsimons, Athanassiadou, Kashem, Madge, et~al.]{ravuri2021skilful}
Suman Ravuri, Karel Lenc, Matthew Willson, Dmitry Kangin, Remi Lam, Piotr Mirowski, Megan Fitzsimons, Maria Athanassiadou, Sheleem Kashem, Sam Madge, et~al.
\newblock Skilful precipitation nowcasting using deep generative models of radar.
\newblock \emph{Nature}, 597\penalty0 (7878):\penalty0 672--677, 2021.

\bibitem[Shi et~al.(2015)Shi, Chen, Wang, Yeung, Wong, and Woo]{shi2015convolutional}
Xingjian Shi, Zhourong Chen, Hao Wang, Dit-Yan Yeung, Wai-Kin Wong, and Wang-chun Woo.
\newblock Convolutional lstm network: A machine learning approach for precipitation nowcasting.
\newblock \emph{Advances in neural information processing systems}, 28, 2015.

\bibitem[Shi et~al.(2017)Shi, Gao, Lausen, Wang, Yeung, Wong, and Woo]{shi2017deep}
Xingjian Shi, Zhihan Gao, Leonard Lausen, Hao Wang, Dit-Yan Yeung, Wai-kin Wong, and Wang-chun Woo.
\newblock Deep learning for precipitation nowcasting: A benchmark and a new model.
\newblock \emph{Advances in neural information processing systems}, 30, 2017.

\bibitem[Sohl-Dickstein et~al.(2015)Sohl-Dickstein, Weiss, Maheswaranathan, and Ganguli]{sohl2015deep}
Jascha Sohl-Dickstein, Eric Weiss, Niru Maheswaranathan, and Surya Ganguli.
\newblock Deep unsupervised learning using nonequilibrium thermodynamics.
\newblock In \emph{International conference on machine learning}, pages 2256--2265. PMLR, 2015.

\bibitem[Song et~al.(2020)Song, Meng, and Ermon]{song2020denoising}
Jiaming Song, Chenlin Meng, and Stefano Ermon.
\newblock Denoising diffusion implicit models.
\newblock In \emph{International Conference on Learning Representations}, 2020.

\bibitem[Tan et~al.(2023{\natexlab{a}})Tan, Gao, Wu, Xu, Xia, Li, and Li]{tan2023temporal}
Cheng Tan, Zhangyang Gao, Lirong Wu, Yongjie Xu, Jun Xia, Siyuan Li, and Stan~Z Li.
\newblock Temporal attention unit: Towards efficient spatiotemporal predictive learning.
\newblock In \emph{Proceedings of the IEEE/CVF Conference on Computer Vision and Pattern Recognition}, pages 18770--18782, 2023{\natexlab{a}}.

\bibitem[Tan et~al.(2023{\natexlab{b}})Tan, Li, Gao, Guan, Wang, Liu, Wu, and Li]{tan2023openstl}
Cheng Tan, Siyuan Li, Zhangyang Gao, Wenfei Guan, Zedong Wang, Zicheng Liu, Lirong Wu, and Stan~Z Li.
\newblock Openstl: A comprehensive benchmark of spatio-temporal predictive learning.
\newblock In \emph{Conference on Neural Information Processing Systems Datasets and Benchmarks Track}, 2023{\natexlab{b}}.

\bibitem[Tulyakov et~al.(2018)Tulyakov, Liu, Yang, and Kautz]{tulyakov2018mocogan}
Sergey Tulyakov, Ming-Yu Liu, Xiaodong Yang, and Jan Kautz.
\newblock Mocogan: Decomposing motion and content for video generation.
\newblock In \emph{Proceedings of the IEEE conference on computer vision and pattern recognition}, pages 1526--1535, 2018.

\bibitem[Veillette et~al.(2020)Veillette, Samsi, and Mattioli]{veillette2020sevir}
Mark Veillette, Siddharth Samsi, and Chris Mattioli.
\newblock Sevir: A storm event imagery dataset for deep learning applications in radar and satellite meteorology.
\newblock \emph{Advances in Neural Information Processing Systems}, 33:\penalty0 22009--22019, 2020.

\bibitem[Voleti et~al.(2022)Voleti, Jolicoeur-Martineau, and Pal]{voleti2022mcvd}
Vikram Voleti, Alexia Jolicoeur-Martineau, and Chris Pal.
\newblock Mcvd-masked conditional video diffusion for prediction, generation, and interpolation.
\newblock \emph{Advances in Neural Information Processing Systems}, 35:\penalty0 23371--23385, 2022.

\bibitem[Wang et~al.(2017)Wang, Long, Wang, Gao, and Yu]{wang2017predrnn}
Yunbo Wang, Mingsheng Long, Jianmin Wang, Zhifeng Gao, and Philip~S Yu.
\newblock Predrnn: Recurrent neural networks for predictive learning using spatiotemporal lstms.
\newblock \emph{Advances in neural information processing systems}, 30, 2017.

\bibitem[Wu et~al.(2021)Wu, Yao, Wang, and Long]{wu2021motionrnn}
Haixu Wu, Zhiyu Yao, Jianmin Wang, and Mingsheng Long.
\newblock Motionrnn: A flexible model for video prediction with spacetime-varying motions.
\newblock In \emph{Proceedings of the IEEE/CVF conference on computer vision and pattern recognition}, pages 15435--15444, 2021.

\bibitem[Yang et~al.(2022)Yang, Srivastava, and Mandt]{yang2022diffusion}
Ruihan Yang, Prakhar Srivastava, and Stephan Mandt.
\newblock Diffusion probabilistic modeling for video generation.
\newblock \emph{arXiv preprint arXiv:2203.09481}, 2022.

\bibitem[Yu et~al.(2023)Yu, Sohn, Kim, and Shin]{yu2023video}
Sihyun Yu, Kihyuk Sohn, Subin Kim, and Jinwoo Shin.
\newblock Video probabilistic diffusion models in projected latent space.
\newblock In \emph{Proceedings of the IEEE/CVF Conference on Computer Vision and Pattern Recognition}, pages 18456--18466, 2023.

\bibitem[Zhang et~al.(2023)Zhang, Long, Chen, Xing, Jin, Jordan, and Wang]{zhang2023skilful}
Yuchen Zhang, Mingsheng Long, Kaiyuan Chen, Lanxiang Xing, Ronghua Jin, Michael~I Jordan, and Jianmin Wang.
\newblock Skilful nowcasting of extreme precipitation with nowcastnet.
\newblock \emph{Nature}, 619\penalty0 (7970):\penalty0 526--532, 2023.

\end{thebibliography}
